\documentclass{article}

\PassOptionsToPackage{numbers}{natbib}

\usepackage[final]{neurips_2020}

\usepackage[utf8]{inputenc} 
\usepackage[T1]{fontenc}    
\usepackage{hyperref}       
\usepackage{amssymb}
\usepackage{url}            
\usepackage{booktabs}       
\usepackage{amsfonts}       
\usepackage{nicefrac}       
\usepackage{microtype}      
\usepackage{xcolor}         
\usepackage{times}
\usepackage{amsmath, amssymb}
\usepackage{mathtools}
\usepackage{latexsym}
\usepackage{todonotes}
\usepackage{comment}
\usepackage{tabto}
\usepackage{soul}
\usepackage{multicol,multirow}
\usepackage{graphicx}
\usepackage{verbatim}
\usepackage{authblk}

\def\BibTeX{{\rm B\kern-.05em{\sc i\kern-.025em b}\kern-.08em
    T\kern-.1667em\lower.7ex\hbox{E}\kern-.125emX}}
    
\usepackage{color, colortbl}    
\usepackage{amssymb}
\usepackage{algorithm}
\usepackage[noend]{algpseudocode}
\usepackage{xcolor}

\author[1]{\textbf{Thierry Desot}}
\author[1]{\textbf{François~Portet}}
\author[1]{\textbf{Michel Vacher}}

\affil[1]{Univ. Grenoble Alpes, CNRS, Inria, Grenoble INP, LIG, 38000 Grenoble, France  \qquad}

\title{End-to-End Spoken Language Understanding: Performance analyses of a voice command task in a low resource setting}
\usepackage{natbib}
\usepackage{graphicx}
\usepackage{soul}
\usepackage{lineno,hyperref}
\usepackage{listings}
\usepackage{tikz}
\usepackage{verbatim}
\usepackage{amsmath}
\usepackage{tabularx}
\usepackage[figuresright]{rotating}
\usetikzlibrary{shapes}
\modulolinenumbers[5]
\usepackage[title,titletoc,page]{appendix}

\begin{document}

\maketitle

\begin{abstract}

Spoken Language Understanding (SLU) is a core task in most human-machine interaction systems 
. With the emergence of smart homes, smart phones and smart speakers, SLU has become a key technology for the industry. In a classical SLU approach, an Automatic Speech Recognition (ASR) module transcribes the speech signal into a textual representation from which a Natural Language Understanding (NLU) module extracts semantic information. Recently 
End-to-End SLU (E2E SLU) based on Deep Neural Networks has gained momentum since it benefits from the joint optimization of the ASR and the NLU parts, hence limiting the cascade of error effect of the pipeline architecture.
However, little is known about the actual linguistic properties used by E2E models to predict concepts and intents from speech input. In this paper, we present a study identifying the signal features and other linguistic properties used by an E2E model to perform the SLU task. The study is carried out in the application domain of a smart home that has to handle non-English (here French) voice commands.

The results show that a good E2E SLU performance does not always require a perfect ASR capability. 
Furthermore, the results show the superior capabilities of the E2E model in handling background noise and syntactic variation compared to the pipeline model. Finally, a finer-grained analysis suggests that the E2E model uses the pitch information of the input signal to identify voice command concepts. The results and methodology outlined in this paper provide a springboard for further analyses of E2E models in speech processing.

\end{abstract}

\section{Introduction}

Spoken Language Understanding (SLU)~\citep{tur:2011} is a crucial task of most modern systems interacting with humans by speech. SLU refers to the ability of a machine to extract semantic information from speech signals in a form which is amenable for further processing including dialogue, voice commands, information retrieval, etc.  

With the emergence of smart assistants in computer, smart phones and smart speakers, SLU is not only a great research area but has also become a key technology for the  industry, as evidenced by several challenges including major companies such as Amazon with its Alexa prize of several million dollars to be distributed to university teams demonstrating ground breaking progress in spoken conversational AI\footnote{\url{https://developer.amazon.com/alexaprize}}.

From a computing perspective, the classical approach to SLU since the early age of spoken language systems \citep{Sears1988,hemphill1990} has been a pipeline of Automatic Speech Recognition (ASR) -- to transcribe speech signal into a textual representation -- which feds a Natural Language Understanding (NLU) module -- to extract semantic labels from the transcription. The main problem of such an  approach is the cascading error effect which shows that any error at the ASR level has a dramatic impact on the NLU part. This is why a large set of different approaches has been proposed to take into account the ASR hypotheses uncertainty within the NLU module to improve the robustness of the whole chain. This `classical' approach is the default one in most industrial applications and is still an active research area \citep{simonnet2017asr}. 

However, since \cite{qian2017exploring}, Neural End-to-End SLU (E2E SLU) has emerged to benefit both from the performing architecture of Deep Neural Networks (DNN) and from the joint optimisation of the ASR and NLU parts. 
Although the classical pipeline SLU model is still competitive, E2E SLU research has shown that the joint optimisation is an efficient way to handle the problem of cascading errors \citep{serdyuk2018towards,desot2019slu,desot2019towards,ghannay2018end}. In particular, it has been shown that perfect ASR transcriptions are \textit{not} necessary to predict intents and concepts \citep{ghannay2018end}. 

However, E2E SLU is more than just a way to perform joint optimisation. Indeed, in Neural End-to-End SLU, contrary to the classical pipeline, the decision stage has a direct access to the acoustic signal (e.g. prosody features). Therefore, an important question is to know \textbf{which signal characteristics (and other linguistic properties) are effectively used by the E2E SLU model}. To the best of our knowledge there has been little research to shed light on this important question.
Therefore, the goal of this study is to perform a comprehensive analysis of the linguistic features and abilities that are better exploited by E2E SLU than pipeline SLU. 
In particular, this paper addresses the following research questions:
\begin{enumerate}
    
    \item \textbf{Can the cascade error effect of the pipeline SLU approach be avoided?} Although, this has been showed in other research, there are still too few research projects 
    to take the answer as granted. Hence we present a comparison between pipeline and E2E SLU approach evaluated in a realistic voice command context for a language other than English: French.
        
    \item \textbf{Is the model effectively exploiting acoustic information to perform concept and intent prediction?} We are not aware of studies having seriously explored this question. 
    We believe that an E2E model accesses the \textit{acoustic} levels to infer concepts and intents directly from speech. By accessing this information the model can avoid the \textit{cascade of errors} introduced by the interaction between the ASR and NLU models in a pipeline SLU method. 
    
    \item \textbf{Would an E2E SLU model be able to be more robust to variations in vocabulary and syntax?} While grammatical robustness seems to be a question only for the NLU task, we are not aware of research having investigated whether an E2E model would present a better ability to process grammatical variation. Hence, we present an acoustic and grammatical performance analysis to assess the ability of E2E SLU models to handle variation at these two levels.

\end{enumerate}

Part of the comparison between pipeline and E2E SLU has been published in \citep{desot2019towards,desot2019slu}. 

However, this paper presents updated experiments and a transfer learning approach which has not been presented yet. Furthermore, the comprehensive acoustic and symbolic performance analysis was never published and constitutes the core of this article.

In this paper, Section~\ref{sec:SLU_state_of_the_art} gives a brief overview of the state-of-the-art dedicated to pipeline and E2E SLU as well as the few studies analysing deeply E2E SLU performances. Our whole approach is described in Section~\ref{sec:method} where we recap the motivations, the baseline pipeline SLU and the target E2E SLU approaches before introducing the evaluation strategy. Section~\ref{sec:Data_test_train_overview} summarises the artificial speech training data generation and describes the held out real speech test set acquired in a real smart home. Section~\ref{sec:Experiments_and_results} presents the results of the experiments with the baseline SLU and E2E SLU trained by transfer learning. It reports superior performances for the E2E SLU model despite its lower ability at the speech transcription task than the pipeline model. Sections~\ref{sec:Acoustic_impact_on_E2E_SLU_prediction} and ~\ref{sec:Symbolic_impact_on_E2E_SLU_prediction} present respectively the analysis at the acoustic level and at the grammatical level. The E2E SLU model has shown to be more robust than the pipeline one at the noise and grammatical variation levels while its ability to benefit from prosodic information is less clear. These findings are discussed in Section~\ref{sec:Discussion} before reaching the conclusion in Section~\ref{sec:Conclusion}.

\section{Related work} \label{sec:SLU_state_of_the_art}

Although there is a rising interest in End-to-End (E2E) SLU that jointly performs ASR and NLU tasks, the E2E models have still not definitely superseded pipeline approaches. 

\subsection{Pipeline SLU} \label{sec:SLU_SOTA_pipeline}

A typical SLU pipeline approach is composed of an ASR and a NLU module. ASR output hypotheses are fed into an NLU model aiming to extract the meaning from the input transcription. The main problem with such an approach is the dependence on the transcription output from the ASR module causing error propagation and reducing the performance of the NLU module.

Hence, to deal with the uncertainty conveyed by the ASR, several methods incorporate the handling of \textit{N best hypotheses}. For instance in \cite{he2003data}, the ASR module (HMM) is followed by an NLU module using a Hidden Vector State Model (HVS) for concept prediction.
A rescoring is applied to the N-best word hypotheses from word lattices as output from the speech recogniser. Parse scores from the semantic parser are then combined with the language model likelihoods.

Another strategy to decrease error propagation is the use of \textit{confidence measures}. These were used by 
\cite{sudoh2006incorporating} for concept prediction, augmenting Japanese ASR transcriptions with concept labels. In their case, a concept label was associated to ASR transcriptions by an SVM model only if confidence measures were above a certain threshold. 
N-best list hypotheses and ASR output confidence measures were also exploited  using weighted voting strategies \citep{zhai2004using}. Since the {n}\textsuperscript{-th} hypothesis transcription contains more errors than the {n-1}\textsuperscript{-th} hypothesis, voting mechanisms were used to improve performance. For instance, a concept was considered correct if it was predicted in more than 30\% of the n-best hypotheses per reference sentence. 

A third strategy is to use \textit{word confusion networks}. For instance 
\cite{hakkani2006beyond} improve the transition between ASR and SLU concept prediction, using word confusion networks obtained from ASR word lattices instead of simply using ASR one-best hypotheses \citep{mangu2000finding}. Word confusion networks provide a compact representation of multiple aligned ASR hypotheses along with word confidence scores. Their transitions are weighted by the acoustic and language model probabilities. 

More recently, acoustic word embeddings for ASR error detection were trained through a convolutional neural network (CNN) based ASR model to detect erroneous words \citep{simonnet2017asr}. This approach was combined with word confusion networks and posterior probabilities as confidence measures, for concept prediction. Output of the ASR model was fed to a conditional random fields (CRF) model and an attention-based RNN NLU model.

In \cite{liu2020jointly}, SLU models using word confusion networks are compared with 1-best hypothesis and N-best lists (N=10) for concept label and value prediction. The ASR posterior probabilities are integrated in a pretrained BERT  based SLU model \citep{devlin-etal-2019-bert}. The word confusion network is fed into the BERT encoder, and integrated into vector representation. The output layer is a concept label and value classifier.  

Finally, another strategy, particularly adapted when aligned labels are missing (e.g., different from an aligned BIO scheme), is to use a \textit{sequence generation} instead of a \textit{sequence labelling} approach. In \citep{desot2019towards,desot2019slu}, \textit{unaligned NLU} data was used to train a BiLSTM seq2seq attention-based model as NLU module. Despite a lower prediction accuracy than aligned models, it provides the flexibility to infer slot labels from imperfect transcriptions and speech with disfluencies.

\subsection{E2E SLU} \label{sec:SLU_SOTA_E2E}

Only recently SLU is conceived as a joint processing of the ASR and NLU tasks which decreases the error propagation between the ASR and NLU modules. 
Furthermore, such a model has \textit{access to the acoustic and prosodic levels}, which can have a positive impact on the performance of SLU. For instance, 
\cite{serdyuk2018towards} trained a sequence-to-sequence (seq2seq) model on clean and noisy speech data to infer intents directly from audio MFCC. Such an approach showed that some prosodic aspects of the speech signal were exploited by the E2E model for intent classification (e.g., question vs imperative voice). 

E2E SLU is also driven by the intuition that recognising speech,\textit{word by word}, is \textit{not} necessary. 

In 
\citep{ghannay2018end}, the Baidu Deep Speech ASR system \citep{hannun2014deep} was trained on transcriptions enriched with concept labels. Eight concepts were injected into the ASR transcriptions as symbolic labels. In order to reduce the importance that the connectionist temporal classification (CTC) cost function  \citep{graves2006connectionist,watanabe2017hybrid,ueno2018acoustic} assigns to each character and to draw more attention to the concept symbols, all character sequences not related to a concept label were replaced by one and the same symbol.

Different from these E2E models that predict intent and concept labels directly from speech, a \textit{transfer learning} technique allows the training of a complete E2E model though sub-tasks (for instance, forcing hidden layer to predict phonemes), thereby providing an easier learning path. Combined with a curriculum learning that presents the easy examples before the more complex ones during training, convergence of the learning algorithms is accelerated \citep{krueger2009flexible}. For instance, in 
\cite{lugosch2019speech}, a transfer learning for intent prediction is applied by training first an ASR model and then adapting it to an SLU task to predict concepts  
that are finally mapped to intents.

In \cite{caubriere2019curriculum,caubriere2020we}, this type of E2E SLU is performed for concept prediction, using the Baidu Deep Speech ASR tool. 
A phase of training an ASR model, is followed by three phases of learning concepts with an increasing complexity. The approach showed a clear gain in performance compared with a classical pipeline approach on the French Media corpus. 

\subsection{E2E SLU analysis} \label{sec:SLU_SOTA_E2E_analysis}

Although recently E2E SLU is emerging as an alternative approach for pipeline SLU systems, we were not able to find research offering an in-depth performance analysis.  
In 
\cite{caubriere2019curriculum}, an \textit{Error analysis} of their E2E SLU system is performed. They show that concept deletion errors are not mainly caused by the ASR capability of the system, but occur as a consequence of a segmentation problem. On top of that, unseen concepts are better predicted using a transfer learning approach. In the study of 
\cite{rao2020speech}, a pipeline SLU system is compared with an E2E SLU model for Amazon Alexa, where the interface between ASR and NLU is a shared 1-best hidden layer. They show that joint ASR and NLU training improves SLU performances for ASR erroneous output transcriptions that impact NLU performances in a pipeline model. 

In 
\cite{denisov2020pretrained}, a pretrained ESPnet ASR model encoder was combined with transformer based pretrained contextual BERT embedding. They analysed the fine-tuning of the E2E SLU \textit{layers}. 
Results indicate that fine-tuning the ASR encoder layers is more beneficial than the NLU layers. This would mean that the acoustic representation must be adapted to the concept extraction task and is thus different from the ASR task.

This brief state of the art shows that handling the cascading error effect has been the main focus of pipeline SLU studies and the main motivation for E2E systems. Despite a gain of performance of E2E SLU and some studies analysing the potential effect of such gain, 
the impact of the E2E SLU model's access to the acoustic level has not been investigated in-depth. To the best of our knowledge, we are not aware of any other study analysing the impact of acoustic features on E2E SLU performances and the robustness of such a model to grammatical mismatch.

\section{Method} \label{sec:method}

Before presenting the overall approach of the paper, 
we recall the research questions mentioned in the introduction. 

The state of the art section showed that one of the main drawbacks of a pipeline SLU is the cascade of errors effect
. Hence, the main objective of previous and recent SLU approaches was to reduce the impact of ASR errors on NLU performances. Our goal in this study is not only to avoid this cascade of errors but also to \textbf{understand what advantages an end-to-end (E2E) SLU approach can offer over a traditional pipeline approach} (this will be detailed in Section \ref{sec:Experiments_and_results}).

Given that the E2E approach extracts semantics directly from the acoustic signal, an interesting research question we address in this paper is \textbf{ whether the E2E model exploits prosodic information to infer intent and concepts from the speech signal} (this will be the subject of Section \ref{sec:Acoustic_impact_on_E2E_SLU_prediction}). 

Another drawback of the classical ASR and NLU systems that can impact the NLU part negatively is the problem of out of vocabulary words (OOV) and unusual syntactic structures. By modelling linguistic phenomena at a finer granularity, we want to verify \textbf{whether an E2E SLU model is more robust to vocabulary and syntactic variations} (Section \ref{sec:Symbolic_impact_on_E2E_SLU_prediction} will detail this study).

This section details the overall approach we have followed to address the above challenges.

 \subsection{Overall approach}

The solution we propose consists in considering the SLU problem as a slot-filling problem applied to the field of voice command in a smart home.

\begin{figure}[!ht]
\centering
\begin{tikzpicture}[node distance=2.5cm, auto, text width=2cm]  
  \node [draw]  (corpus)  {Corpus acquisition};  
  \node [draw,right of=corpus]  (SLU) {Definition of models};  
  \node [draw,right of=SLU]  (SLU_app) {Model training};
  \node [draw,right of=SLU_app] (eval) {Evaluation \& analysis};
  \path [draw,->] 
                (corpus) -- (SLU)  
                (SLU) -- (SLU_app) 
                (SLU_app) -- (eval) ;  
    
\end{tikzpicture}
\caption{Overview of SLU development steps\label{fig:demarche}}
\end{figure}
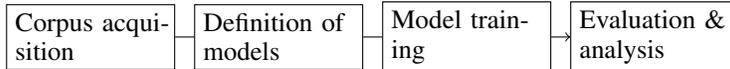

Figure~\ref{fig:demarche} describes the steps of the approach. 
The first step was to collect a 
\textit{test corpus} to evaluate the SLU approaches on realistic data. To solve the lack of \emph{training} data, we have chosen an expert based artificial corpus generation approach. In order to reproduce a realistic situation, no data from the test corpus was used to train the models. 

Then, we defined two baseline systems: the pipeline SLU and E2E SLU. 
\textit{The pipeline SLU} is a combination of 2 state of the art ASR and NLU modules. \textit{The End-to-end SLU model} is a pyramidal RNN multi-task model that combines a CTC cost function and an attention based encoder and decoder . Theoretically, this type of model is capable of handling OOV words, it is therefore interesting to study whether the interaction between attention and CTC can strengthen the robustness of such a model on test data with high \textit{linguistic variability}. The SLU objective and the two SLU baselines are further introduced in the following Section~\ref{sec:SLU_approach}.  

Once the models defined, they were trained on the generated corpus. The pipeline training is detailed in Section~\ref{subsec:pipeline_SLU} and
the E2E training in Section~\ref{subsec:E2E_SLU}. 
Finally, once the models have been trained for the task, we performed various evaluations to assess to which extent the results are correlated to external factors such as noise condition, gender, pitch variation, syntactic complexity, etc. The correlation measures used for the study are introduced in  Section~\ref{sec:analysis_measures} while the experiments are detailed in Sections~\ref{sec:Acoustic_impact_on_E2E_SLU_prediction} and~\ref{sec:Symbolic_impact_on_E2E_SLU_prediction}.

\subsection{SLU approach} \label{sec:SLU_approach}
The target of our SLU experiments, are commands without linguistic context and with one intent per utterance. 
The notion of intent is close to that of a \textit{speech act} which is the speaker's communicative activity in which an utterance produces an effect on its interlocutor \citep{crystal:2011}. In this study, the intent is the type of act addressed to the home automation system. For example, the statement ``Allume la lumière'' (\emph{turn on the light}) conveys the intent to the smart home to change the state of an object. To characterise the voice command, it is also necessary to identify its \textit{concepts} or \textit{slots} representing the most important information (i.e., entities and actions). This process is called \textit{slot-filling} \citep{tur:2011}. 
Figure~\ref{fig:schema_architects_SLU_SEQ_SLU_E2E} shows the two ways of extracting intent and slots we followed in this paper. In the pipeline case (on the top of the figure), the input utterance is first transcribed by the ASR module to be analysed by an NLU module that generates a sequence of  
\texttt{concepts} 
that are supposed to be found in the speech input. In the example, the intent is \texttt{set\_device} while the concepts are the action with value \texttt{turn\_on} and the device with value \texttt{light}. In the E2E case (on the bottom of the figure), the SLU target is seen as an enriched transcription task. The SLU model is trained to surround the transcribed words with specific characters such as \string^ to delimit an action or \} to delimit a device. Intents are classified using a special character at the beginning and end of a transcription (here @ means the \texttt{set\_device} intent).   

\tikzstyle{rect} = [draw, rectangle, fill=white!20, text width=5em, text centered, minimum height=2em]
\begin{figure}[!ht]
\begin{tikzpicture}[node distance=1cm, auto] 
\node [rect,draw=none] (step1) {{\small ``Allume la lumière*''}};
\node [right of=step1, node distance=2.5cm] (uk_1) {{\footnotesize \emph{*Turn on the light}}};
\node [rect, below right of=step1, node distance=2.5cm] (step3) {E2E SLU};
\node [rect, above right of=step1, node distance=2.5cm] (step4) {ASR};
\node [rect, draw=none,right of=step4, node distance=2.5cm] (step5) {{\small``Allume les lumières**''}};
\node [below right of=step5, node distance=2cm] (uk_2) {{\footnotesize \emph{**Turn on the lights}}};
\node [rect, right of=step5, node distance=2.75cm] (step6) {NLU};
\node [rect,draw=none, right of=step6, node distance=3cm,  text width=3cm] (step7) {{\small intent[set\_device], action[turn\_on], device[light]}};

\node [rect,draw=none, right of=step3, node distance=8.3cm,  text width=3cm] (step8) {{\small@ \string^allume\string^ \}la lumière\} @}};

\path [draw ,->] (step1)  -- (step4);
\path [draw ,->] (step1)  -- (step3);
\path [draw ,->] (step4)  -- (step5);
\path [draw ,->] (step5)  -- (step6);
\path [draw ,->] (step6)  -- (step7);
\path [draw ,->] (step3)  -- (step8);
\end{tikzpicture}
\caption{Comparison of pipeline and E2E SLU tasks}
\label{fig:schema_architects_SLU_SEQ_SLU_E2E}
\end{figure}
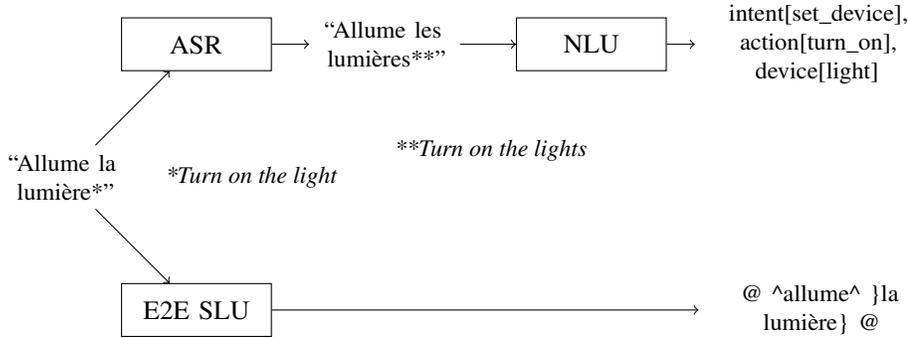

\subsection{Baseline pipeline SLU} \label{Baseline_pipeline_slu}

As in \citep{desot2019towards}, the ASR component of our pipeline SLU is the Kaldi tool, nnet2 version. This neural-network ASR training framework allows training with large amounts of data using multiple GPUs or multi-core machines. It uses speaker adapted features from the GMM (Gaussian Mixture Model) system, so a first pass of GMM decoding and adaptation is required \citep{povey2011kaldi,povey2014parallel}.

Mel Frequency Cepstral Coefficients (MFCC) were used as input features. Kaldi also allows using several adaptation methods of the acoustic models to the speaker, such as {\it Maximum Likelihood Linear Regression} (MLLR) \citep{Leggetter1995}, {\it Constrained Maximum Likelihood Linear Regression} (fMLLR) \citep{digalakis1996speaker} and {\it Speaker Adaptive Training} (SAT) \citep{anastasakos1996compact}. As the ASR component has to interact with the NLU module in a pipeline system in the real time setting of a smart home, the nnet2 \textit{online} version was also used.

Regarding the NLU module, we approach it as a \textit{sequence generation task} with \emph{unaligned} data. The SLU task is seen as a translation problem where the input must be abstracted to generate output intent classes and concept labels. For that reason, the NLU module was a seq2seq bi-directional LSTM encoder and decoder attention-based model. 
This was our strategy to decrease errors of our baseline pipeline SLU model, due to the imperfect transcription output of the ASR component that impacts the NLU. Using an unaligned approach the model should learn to associate several words to one slot label without aligned data. Furthermore, classical BIO alignment can not be assumed for pipeline and E2E SLU when input data consists of spontaneous speech with disfluencies that often cause ASR deletion and insertion errors. Hence, a robust NLU sub-part is needed that can handle those and that can be trained with unaligned labels. In \cite{mishakova2019learning}, we showed that such an NLU approach is competitive with state of the art \emph{aligned} NLU CRF models \citep{Jeong2008} and also with DNN-based models \citep{Mesnil2015,Bapna2017,Liu2016,Huang2017} that deal with the NLU problem as a \textit{sequence labelling task}.

\subsection{End-to-end SLU}

\begin{figure}[t]
\centering
\includegraphics[width=0.75\textwidth]{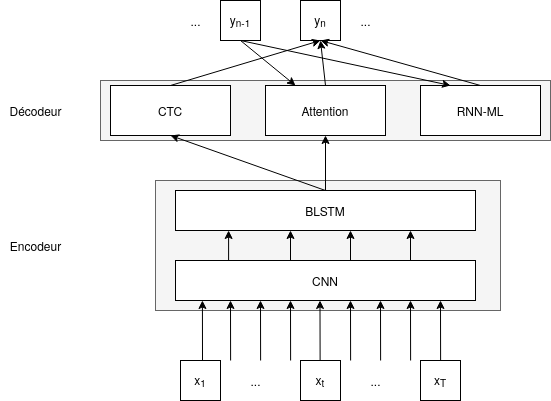} 
\caption{ESPnet architecture with the different training/inference strategies (CTC, attention, ML RNN)}
\label{fig:ESPnet_CTC_attention_ML-RNN}
\end{figure}

The E2E approach as outlined in \cite{desot2019slu} was based on the ESPnet ASR toolkit \citep{watanabe2018espnet}. It integrates the Kaldi data preparation, extracts Mel filter-bank features, and combines Chainer and PyTorch deep learning tools \citep{tokui2015chainer,paszke2017automatic}. The encoder consists of a very deep convolutional neural network (VGG) followed by six bidirectional pyramidal subsampling bi-LSTM layers. Figure~\ref{fig:ESPnet_CTC_attention_ML-RNN} includes an overview of the ESPnet architecture.

Convolutional neural networks (CNNs) have achieved great success in image recognition \citep{cho2018multilingual}. In the context of ASR, CNNs are usually used as feature extractors, while the HMM part is typically replaced by RNNs that provide a distribution over sequences directly \citep{zhang2016towards}. 
The success of using CNNs in ASR tasks, can be  attributed to the use of local filtering and maxpooling in the CNN architecture. This combination 
turns out to be a better strategy than a GMM model that represents the entire frequency spectrum as a whole. Another benefit is better robustness against ambient noise. In order to locally apply filtering, a frequency scale is needed that can be divided into a number of local bands. 
Therefore, MFCC features are not suitable because of DCT-based (Discrete Cosine Transform) decorrelation transform. Indeed the Gaussian mixture model needed decorrelated input feature dimensions while CNNs can  benefit from correlated inputs. For that reason, filter-bank features are more fit for local filtering using CNNs. Max pooling and filtering is also an alternative for SAT (Speaker Adaptive Training) and MLLR (Maximum Likelihood Linear Regression) that transform speech features into a canonical speaker space \citep{abdel2012applying}. 

In ESPNet, mapping from acoustic features to character sequences is performed by a \textit{hybrid} multitask learning that combines CTC \citep{amodei2016deep,graves2006connectionist} and attention \citep{bahdanau2014neural}. The attention mechanism allows a more flexible alignment, which focuses on the important features and character sequences whereas the ASR alignment is monotonic. A trade-off hybrid CTC and attention-based approach finds a balance between attention and CTC. 

For moving the ESPNet model from a transcription objective to a SLU objective, as described in Figure \ref{fig:schema_architects_SLU_SEQ_SLU_E2E} the transcriptions were enriched with symbols representing intent classes and concept labels. This strategy has been previously used with success by~ \cite{ghannay2018end,desot2019slu, desot2019towards}.

\subsection{Analysis measures}\label{sec:analysis_measures}

An E2E SLU approach has access to the \textit{acoustic} and \textit{prosodic} information of the input signal. Therefore, in the analysis we intend to assess to which extent the model is capable of exploiting para-linguistic information to infer semantic information. 
Table \ref{tab:Evaluation_model_acoust_pros_symb} gives an overview of the analysis levels that we consider in the study.

\begin{table}[!bh]
\caption{
Analysis levels of ASR, pipeline and E2E SLU on the VocADom@A4H test set}
\label{tab:Evaluation_model_acoust_pros_symb}
\centering
\begin{footnotesize}
\begin{tabular}{|l|ccccccc|ccccc|ccccc|cc|}

\hline
\textbf{Analysis}  &\multicolumn{12}{c|}{\textbf{Acoustic}} &    \multicolumn{7}{c|}{\textbf{Symbolic}} \\
\textbf{level}  &\multicolumn{12}{c|}{\textbf{}} &    \multicolumn{7}{c|}{\textbf{}} \\
\hline
 &\multicolumn{7}{c|}{\textbf{Source}} & \multicolumn{5}{c|}{\textbf{Prosody}}  & \multicolumn{5}{c|}{\textbf{Lexical}} & \multicolumn{2}{c|}{\textbf{Syntax}}\\
\hline
\textbf{Analysis:}  &\multicolumn{4}{c|}{\textbf{Noise}} & \multicolumn{3}{c|}{\textbf{Gender}} & \multicolumn{2}{c}{\textbf{Avg. F0}} & \multicolumn{3}{c|}{\textbf{}} & \multicolumn{5}{c|}{\textbf{OOV}} & \multicolumn{2}{c|}{\textbf{Variation}}  \\
\hline
\textbf{Hypothesis:} & \multicolumn{19}{l|}{(1) 
Does the E2E model benefit from \textbf{prosodic, acoustic information}?
} \\
\textbf{} & \multicolumn{19}{l|}{(2) 
Is the E2E model more robust to \textbf{lexical} and \textbf{syntactic} variations?} \\
\hline

\end{tabular}
\end{footnotesize}
\end{table}

The acoustic analysis was devoted to two main features~: the robustness to the variability of the source (here, the background noise and gender) and the capability of the model to exploit acoustic features. The robustness to OOV and syntactic variability was studied under the term 'symbolic' in order to distinguish it from the purely acoustic part of the study. For each study, the analysis consisted in generating or extracting specific stimuli to feed models, then measuring the \emph{performance} of each model at various levels (intent, slot-filling, speech recognition) and assessing their \emph{correlations} with the acoustic or symbolic features of the input stimuli. We describe below which performance and correlation measures we used in the study.

For assessing the intent prediction performance, since the task consists in choosing for each utterance one possible intent among a restricted set of classes, we therefore used classic measures in classification, recall $= \frac{TP}{TP+FN}$ and precision $= \frac{TP}{TP+FP}$, where TP is {\it True Positive}, FP is {\it False Positive}, and FN {\it False Negative}. \textit{Precision} expresses the proportion of correctly predicted intents in the set of \textit{predictions}, whereas \textit{recall} expresses the rate of correct predictions among the set of instances to be predicted. The F-measure or F1 score provides a single score that balances both precision and recall and is calculated as follows,

\begin{equation}
F1 = 2 \times \dfrac{\text{Precision} \times \text{Recall}}{\text{Precision} + \text{Recall}}
\end{equation}

\begin{figure}[ht]
\centering
\includegraphics[width=0.6\textwidth]{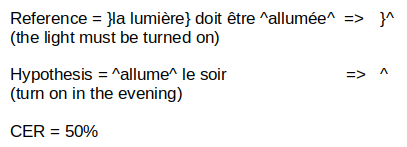} 
\caption{E2E SLU - calculation of the concept error rate that is symbolically represented}
\label{fig:CER_E2E_example}
\end{figure}

For slot-filling performance 
we used concept error rate (CER) which is defined as the ratio of the sum of deleted, inserted and confused concepts w.r.t. a Levenshtein alignment for a given reference concept string \citep{hahn2008comparison}. In this paper, we calculated the CER in a similar way, but we did not take the label sequence order into account since we used a generative approach. Hence, a reference sequence {\tt action, device} provides the same information as a hypothesis such as {\tt device, action}.
As for the E2E approach, transcriptions are enriched with symbols that represent concepts and intents. We only consider the symbol sequences of reference and hypothesis concepts. Figure~\ref{fig:CER_E2E_example} shows how the labels are extracted from the outputs to compute the  CER. In this example, the hypothesis transcription shows an erroneous deletion of the symbol \verb+}+  and obtains thus a CER score of 50\%.

For the ASR level, performances were evaluated using the WER. The WER is defined as the sum of \textit{insertions} ($I$), \textit{deletions} ($D$) and \textit{substitutions} ($S$) of words compared to the number of words $N$ in a reference transcription that has manually been verified. The alignment between the reference and hypothesis transcription is obtained by dynamic programming in order to find the alignment leading to the minimal WER. The WER is calculated as follows~:

\begin{equation}
\text{WER} = \dfrac{I + S + D}{N} \times 100
\end{equation}

As a final evaluation step, our main strategy is to compute the correlation of the resulting SLU and ASR performances with respect to the characteristics of the input stimuli. 
We measure \textit{Pearson} and \textit{Spearman's rank} correlations between performances (CER, WER, F1) and input features (e.g. \textit{energy}).

Pearson's correlation coefficient $r$ measures the strength of the association between two variables $x$ and $y$ assuming a normal distribution of values. Pearson's correlation is calculated as follows, where $n$ denotes the number of elements~:

\begin{equation}
r = \frac{\sum_{i=1}^{n} \left(x_{i} - \overline{x}   \right) \left( y_{i} - \overline{y}  \right)}
{   \sum_{i=1}^{n}\sqrt{\left(x_{i} - \overline{x}   \right)^2}  \sum_{i=1}^{n}\sqrt{\left(y_{i} - \overline{y}   \right)^2} } 
\end{equation}

$r$ varies between -1 and 1. There is no correlation when $r$ is equal to 0. 1 indicates a strong correlation and -1 a strong negative correlation.

\textit{Spearman's rank correlation}, $r_{s}$ , measures correlations between \textit{ranked variables} and is calculated as follows where $n$ denotes the number of elements~: 

\begin{equation}
r_{s} = 1 - \frac{6\sum_{i=1}^{n} \left(x_{i} - y_{i} \right)^2  }{n(n^2-1)}
\end{equation}

In order to determine if the resulting correlation coefficient is significant, the $p$-value is often used. It is calculated in hypothesis testing to determine whether or not to reject a null hypothesis. The $p$-value for the Pearson or Spearman correlation coefficient ($coef$) uses the distribution law $t$, 

\begin{equation}
t = r \sqrt{\frac{n - 2}{1 - coef}} 
\end{equation}

We reject the null hypothesis, $H_{0}: coef = 0$, if the $p$-value is less than 0.05 ($p < 0.05$). The $p$-value or hypothesis  $H_{1}: coef \neq 0$ (two tails test), is computed as follows~:

\begin{equation}
p = 2 * P(T > |t|), 
\end{equation}

where $P$ denotes the probability and $T$ follows a distribution $t$ with $n - 2$ degrees of freedom. We distinguish between correlations for which $p<0.05$ that we mark with $^*$
and correlations for which $p<0.01$ that we denote with $^{**}$.

In order to measure and compare the robustness of the pipeline and E2E SLU models to lexical and grammatical variations (see Section \ref{sec:Symbolic_impact_on_E2E_SLU_prediction}), we verified the impact of an increased OOV (Out Of Vocabulary words) on SLU performances by gradually replacing domain specific vocabulary by synonyms that do \textit{not} occur in the training set.
We also measured the impact of the syntactic variability in the speech of our target users by predicting concepts and intents of the test data where we inserted syntactic structures and disfluencies that hardly occurred in the training set (Table  \ref{tab:Evaluation_model_acoust_pros_symb}).
Our hypothesis is that the learning of the E2E SLU model, that combines the CTC and attention mechanisms, can enhance the robustness of the E2E model to \textit{linguistic variability}.

\section{Data} \label{sec:Data_test_train_overview}

To acquire target corpora in sufficient size to train deep neural network models, we used a combination of realistic data, synthetic data and out-of-domain data. 

Realistic data was acquired within a real smart home with naive users, using a Wizard-Of-Oz strategy to acquire diverse and contextualised voice commands. Indeed, since our target users are senior adults, they tend to deviate from a too strict grammar \citep{Moller2008,Takahashi2003,Vacher2015}, hence the need for a realistic corpus that accounts for the rich set of possible sentences and pronunciations. This corpus is quickly introduced in section~\ref{sec:test_data} and was made available to the community \citep{portet2019context}.

Acquiring such kind of realistic corpus is highly time consuming and leads to an amount of data far too small for machine learning. We tackled this problem by automatically generating a domain-specific \textit{synthetic speech training} corpus using Natural Language Generation controlled by the semantic space of the smart home. This synthetic generation is described Section~\ref{sec:synthdata}. 

Finally, since the artificial generation presents a good semantic coverage but a poor diversity, we also collected other out-of-domain corpora that can be used either to enrich the training data or to perform transfer learning. This collection of corpora is listed in Section~\ref{sec:hand_out_datasets}.

\subsection{VocADom@A4H test data}\label{sec:test_data}

The VocADom@A4H corpus \citep{portet2019context,desot2018towards} includes about twelve hours of speech data and was acquired in realistic conditions in the Amiqual4Home smart home\footnote{https://amiqual4home.inria.fr}. Eleven participants uttered voice commands while performing activities of daily living for about one hour in different rooms including a kitchen, a living room, a bedroom and a bathroom. Out-of-sight experimenters reacted to participants' voice commands following a wizard-of-Oz strategy to add naturalness to the corpus. Furthermore, experimenters were also present in the home to act as visitors of the participant. At the end, the corpus consists of a mixture of spontaneous and read voice commands, with lexical and syntactic variation. Some of the utterances were recorded with background noise (use of vacuum cleaner, radio, tv etc.).

\begin{figure}[ht]
\centering
\includegraphics[width=1\textwidth]{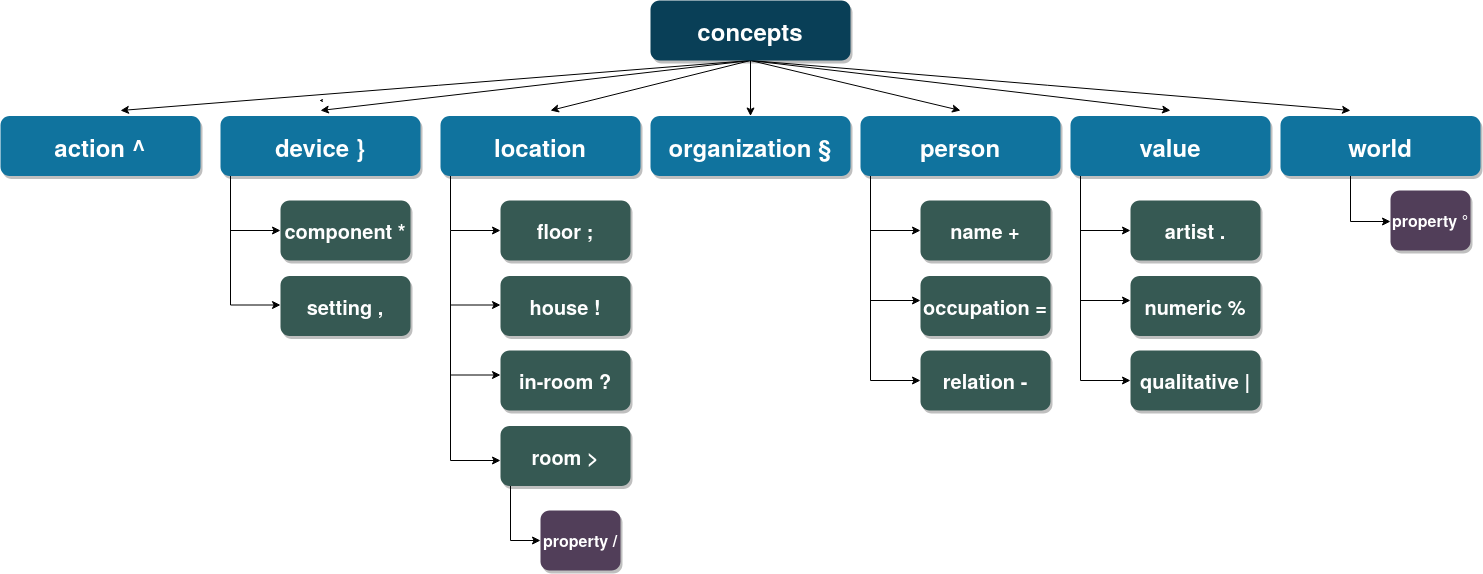} 
\caption{Tree structure of concepts with symbolic representation}
\label{fig:slots_tree}
\end{figure}

Each voice command was prefixed with a keyword (chosen by the participant in a list) to activate the smart home (e.g., ``Minouche, lower the blind of the bedroom'' were `Minouche' was one of the possible keywords). The participants' and experimenters' speech was semi-automatically transcribed and then corrected manually. For SLU, data was manually annotated with intent classes and slot labels whose semantics were defined in accordance with the smart home capabilities. A description of the semantic labels for the slot is provided in Figure~\ref{fig:slots_tree}. 

At the end of the annotation process, 6,747 utterances constitute the dataset. As shown in Table~\ref{tab:corpora_test}, it consists of voice commands (\textit{intents(1)}), and utterances other than voice commands  (\textit{none intents(2)}). This \textit{realistic} corpus is the held out test set used for all our SLU experiments. It is freely available for research purpose at \href{http://vocadom.imag.fr}{vocadom.imag.fr}. For more information about this corpus, the reader is referred to \citep{portet2019context}.

\begin{table*}[t]
\caption{VocADom@A4H test data overview}
\label{tab:corpora_test}
\setlength{\tabcolsep}{3pt}
\centering
\begin{tabular}{|l|c|c|c|c|}

\hline
\textbf{VocADom@A4H}~&\textbf{utterances}~&~\textbf{words}~&~\textbf{intents}~&\textbf{slot labels}\\
\hline
intents(1)& 2612 & 430 & 7 & 14 \\
none intents(2) & 4135 & 1326 & 1 & - \\
\hline
complete(3) & 6747 & 1462 & 8 & 14 \\
\hline
\end{tabular}
\end{table*}

\subsection{Automatic artificial training data generation}\label{sec:synthdata}
To gather a large amount of data with a broad coverage, we used an expert-based Natural Language Generation (NLG) approach \citep{Gatt2018}. An NLG approach is more easily controlled compared to a constrained RNN language model for data augmentation \citep{Hou2018}. The generation was performed in two phases:
\begin{enumerate}
    \item Voice command utterances were generated as text and semantically annotated at the same time.
    \item Generated textual voice commands were fed to a speech synthesizer to provide a complete corpus of speech annotated with semantic information.
\end{enumerate}
    
The NLG system was based on the NLTK python library feature-based context free grammar (FCFG) \citep{bird2009natural} allowing for sentence generation, and for features (i.e. slot information) to be attached to the final output sentences.  The grammar defines intents as a composition of their possible constituents, with constraints on generation. 

The semantic space consisted of four general intent classes:\begin{itemize} \item \texttt{contact} which allows a user to place a call; \item \texttt{set} to make changes to the state of objects in the smart home; \item \texttt{get} to query the state of objects as well as properties of the world at large; \item and \texttt{check} to check the state of an object. Slot labels are divided into eight categories:
the \texttt{action} to perform, the \texttt{device} to act on, the \texttt{location} of the device or action, the \texttt{person} or \texttt{organization} to be contacted, a device \texttt{component}, a device \texttt{setting} and the \texttt{property} of a location, device, or world.\end{itemize}

Syntactical variation was also part of the grammar design.  
Similar to the test set, each voice command includes a keyword to activate the smart home. 
Maximising all combinations of semantic labels that result in meaningful utterances, the grammar generated about 77k phrases. These generated sentences were automatically annotated with 17 different concepts and eight different intent classes, based on the general  categories defined by our semantic space. Table \ref{tab:intents_overview} includes an overview of intents. An overview of concepts is presented in Figure~\ref{fig:slots_tree}, while Table~\ref{tab:intents_overview} provides a comparison of the number of utterances per intent  between the artificial and realistic corpus. The generation process and its evaluation is detailed 
in \citep{desot2018towards}.

\begin{table}[ht]
\caption{Distribution of utterances broken down by intent for the Artificial (Artif.) train and VocADom@A4H (Real.) test corpus with examples}
\label{tab:intents_overview}
\centering
\begin{tabular}{|p{0.3\linewidth}p{0.07\linewidth}p{0.3\linewidth}p{0.07\linewidth} p{0.07\linewidth}|}
\hline
\textbf{Intent} & \textbf{Symb.} & \textbf{Example}  & \textbf{\#}  &  \\
 &  & \textbf{utterances} & \textbf{Artif.} &  \textbf{Real.} \\
\hline
{\tt Check\_device} & \# & {\it minouche is the window open~?}  & 2754 & 284\\
{\tt Contact} & [ & {\it vocadom call a doctor} & 567 & 114\\
{\tt Get\_room\_property} & \{ & {\it berenio what's the temperature~?}  & 9 & 3\\
{\tt Get\_world\_property} & ] & {\it ulysse what's the time~?}  & 9 & 3\\
{\tt None} & & {\it the window is open}  & - & 4135\\
{\tt Set\_device} & @ & {\it hestia lower the blinds}  & 63,288 & 2178\\
{\tt Set\_device\_property} & \_ & {\it ichefix decrease the TV volume}  & 7290 & 9 \\
{\tt Set\_room\_property} & \& & {\it chanticou decrease the temperature}  & 3564 & 21 \\
\hline
\end{tabular}
\end{table}

The semantic annotation part of the synthetic corpus was generated in two versions: one of for the pipeline and one for the E2E approach. Table~\ref{tab:unaligned_corpora_format} provides examples of these two formats. For the E2E SLU approach the artificial corpus transcriptions are enriched with intent class and slot label symbols \citep{desot2019towards,desot2019slu}. A similar approach was applied in \citep{ghannay2018end}, however our transcriptions are enriched with \textit{both} intent and slot label symbols.

\begin{table*}[t]
\caption{Artificial corpus format for the generative approach (pipeline) and the enriched transcriptions approach (E2E)}
\label{tab:unaligned_corpora_format}
\setlength{\tabcolsep}{10pt}
\centering

\begin{tabular}{|l|c|}
\hline
{\bf Format for generative NLU}\\
\hspace{16 mm}{\bf Pipeline SLU} (``vocadom close the door")\\
\verb|(Source) vocadom ferme la porte|\\
\verb|(Target) intent[set_device], action[close], device[door]|\\
\hline
{\bf Format for symbolically enriched transcription}\\
\hspace{16 mm}{\bf E2E SLU} (``vocadom switch on the light")\\
\verb|(Source + Target labels injected)|\\
\verb|@ VocADom                ^allume^         }la lumière} @|\\
\verb|SET_DEVICE| intent class symbol  \verb|@|/  
\verb|Action| slot symbol \verb|^| /  \\
\verb|Device| slot symbol \verb|}|\\

\hline
\end{tabular}
\end{table*}

As an SLU system extracts slot labels and intent classes from speech, we used a speech synthesizer to generate spoken utterances for the 77k artificial sentences, using the open source Ubuntu SVOX~  \footnote{https://launchpad.net/ubuntu/+source/svox}
female French voice~\footnote{https://doc.ubuntu-fr.org/svoxpico}.

\subsection{Collection of realistic data sets}\label{sec:hand_out_datasets}

The artificial corpus, though of great semantic coverage, does not cover the diversity of speech that can be found in the test data. For instance, Table~\ref{tab:corpora_test} reports that \texttt{none} is the majority intent class in the VocADom@A4H test set. Furthermore, the artificial corpus contains only artificial speech produced by one synthetic voice. In order to increase the number of \texttt{none} intents in the artificial training data as well as to add voice diversity, the ESLO2 corpus \citep{serpollet2007large} of conversational French speech was added to the training set. Sentences unrelated to voice command intent were extracted (i.e. \texttt{none} intent) and manually filtered. Only out of domain utterances were kept for collecting \texttt{none} intent training data. Furthermore, similarly to the VocADom@A4H corpus, it contains frequent disfluencies. 
The small domain specific SWEET-HOME corpus, with distant voice commands \citep{vacher2014b} was also added to the training data. 

Table~\ref{tab:corpora} includes an overview of the complete SLU training set consisting of the artificial corpus and the SWEET-HOME and ESLO2 corpora.
The perplexity (\textit{perpl.}) and OOV with respect to the test set are provided using a 3-gram language model learned on each corpus. It can be seen that the vocabulary of the artificial corpus is quite poor despite a good semantic coverage. Because of this small vocabulary and the strict syntactic pattern the perplexity and OOV stays high. By contrast, the SWEET-HOME corpus has a relatively low perplexity even with such a small amount of data. Finally, the large vocabulary of ESLO implies the smallest OOV rate.

\begin{table*}[t]
\caption{Comparison of SLU training and test data (OOV = test set words not seen in training data)}
\label{tab:corpora}
\setlength{\tabcolsep}{3pt}
\centering
\begin{tabular}{|l|c|c|c|c|c|c|c|l|}

\hline
\textbf{training}~&\textbf{utterances}~&~\textbf{words}~&~\textbf{intents}~&\textbf{slots}~&\textbf{perpl.}~&\textbf{OOV}~&\textbf{speech}\\
\textbf{set}~&\textbf{}~&~\textbf{}~&~\textbf{}~&\textbf{}~&\textbf{}~&\textbf{}~&\textbf{(hours)}\\
\hline
Artif. & 77,481 & 187 & 7 & 17  & 124.41 & 307 & 81.25\\
Sweet-Home & 1412 & 480 & 6 & 7  & 49.33 & 343 & 2.5\\
Eslo2 & 161,699 & 29,149 & 1 & - &  151.90 & 211 & 126\\
\hline
Total & 240,592 & 30,821 & 8 & 17   & 372.06 & 235 & 209.75
\\
\hline
\end{tabular}
\end{table*}

\section{Training of the pipeline and E2E SLU models: impact of the cascade error effect} \label{sec:Experiments_and_results} 

\subsection{Baseline pipeline SLU} \label{subsec:pipeline_SLU}

The ASR and NLU modules of the pipeline SLU were trained separately.
For the ASR module, a large acoustic model was trained using 472.65 hours of \textit{Real Speech} data from the French corpora ESTER1 \citep{galliano2005ester} \index{corpus!ESTER1} and ESTER2 \citep{galliano2009ester}, \index{corpus!ESTER2} REPERE \index{corpus!REPERE} \citep{giraudel2012repere}, ETAPE \citep{gravier2012etape}, \index{corpus!ETAPE} BREF120 \citep{tan2006french}, \index{corpus!BREF120} AD \citep{vacher2008preliminary}, \index{corpus!AD} SWEET-HOME \citep{vacher2014b}, \index{corpus!SWEET-HOME} CIRDO \citep{vacher2016cirdo} \index{corpus!CIRDO} and the corpus of spontaneous speech ESLO2 \citep{serpollet2007large}, and also the small domain specific SWEET-HOME corpus (Table \ref{tab:corpus_modacoust_Kaldi}).

For Kaldi NNET2 (Section \ref{Baseline_pipeline_slu}), an architecture of 4 hidden layers with 1024 hidden units, a Softmax layer with 4748 units and SGD (Stochastic Gradient Descent) was used. The learning rate started at 0.01 and ended at 0.0001, with a batch size of 128. The total training lasted for 253 hours, using 1 GPU GeForce GTX TITAN Black. Acoustic features were 13-dimensional MFCC features. For further detail, the reader is refereed  to \citep{desot2019slu,desot2019towards,desot2020corpus}.

\begin{table}[t]
\caption{French corpora used to train an acoustic model using the ASR tool Kaldi
}
\label{tab:corpus_modacoust_Kaldi}
\setlength{\tabcolsep}{3pt}
\centering
\begin{tabular}{|l|c|c|c|}

\hline
\textbf{Corpus}~&~\textbf{\# hours of speech} \\
\hline
ESTER1 & 100 \\
ESTER2 & 100 \\
REPERE & 60\\
ETAPE & 30\\
BREF120 & 51.50\\
AD & 0.5\\
SWEET-HOME & 2.5\\
CIRDO & 2\\
ESLO2 & 126\\
\hline
Total & 472.65\\
\hline
\end{tabular}
\end{table}

For the seq2seq NLU module described in Section~\ref{subsec:pipeline_SLU}, input words were first passed to a 300-unit embedding layer. The encoder and decoder were each a single layer of 500 units. Adam optimiser was used with a batch size of 10, using gradient clipping at a norm of 2.0. Dropout was set to 0.2 and training continued for 10,000 steps with a learning rate of 0.0001. Input sequence length was set to 50 and output sequence length to 20, with a beam search of size 4.
The training data was the combined semantically annotated data sets: artificial, SWEET-HOME and the filtered ESLO2 utterances without intent (Table~\ref{tab:corpora}). 

Once ASR and NLU models trained, inference with the pipeline approach just consisted in feeding the best ASR transcription  generated using Kaldi NNET2
to the seq2seq NLU module. 
Table \ref{tab:eval_SLU_tab} shows the differences in performance between the   pipeline SLU approach (Pipeline SLU) and the NLU model that are larger for slot predictions as compared to intent predictions.

\begin{table}[th]
\caption{Pipeline and E2E SLU performances, \% F1-score - Concept Error Rate - WER on VocADom@A4H. $\dagger$ with ESPNet as ASR}
\label{tab:eval_SLU_tab}
\setlength{\tabcolsep}{4pt}
\centering
\begin{tabular}{|l|lcccc|}

\hline

\textbf{Model}  ~&~ \textbf{Hours} ~&~ \textbf{(\%) TTS}  ~&~ \textbf{Intent} ~&~ \textbf{Slot} ~&~ \textbf{WER} \\
  & \textbf{of speech} ~&~ \textbf{in train} ~&~ \textbf{F1-score} ~&~ \textbf{CER} ~&~ \\ \hline
\textbf{Pipeline:}  & & &      &     & \\
ASR  & 472.65 & 0.00 &  -    &  -   & 22.92 \\
NLU  & - & - &  85.51    &  33.78   & -\\
SLU & 472.65 & 0.00 & \bf{84.21}     &     \bf{36.24} & -\\

\hline
\textbf{E2E:} & & &  &    & \\
ASR & 553.90 & 14.67 & - & - & 46.50 \\
SLU & 553.90 & 14.67 &  47.31     &     51.87 & - \\
\hline
\end{tabular}
\end{table}

\subsection{E2E SLU with transfer learning} \label{subsec:E2E_SLU}
For the E2E experiments, we used ESPnet default settings \citep{desot2019slu} in order to train on speech data, with slots and intents symbolically injected in the transcriptions. Since the training was end-to-end, the training data was composed of the ASR training data (472.65 hrs) plus the NLU training data (81,25) were the artificial dataset was synthesised (cf. Section~\ref{sec:Data_test_train_overview}). This represents 553.9 hours of training data.

Table~\ref{tab:eval_SLU_tab} reports the results on the VocADom@A4H test set (\textit{E2E, ASR}). For the Pipeline and E2E ASR task, a far worse WER is exhibited for the E2E model. Similarly, for the SLU task, the E2E model exhibits far worse performances for CER and F1. 

The large amount of data for the E2E learning, far from allowing generalisation, mainly led to unbalanced learning. In the training data, the open domain was too large to allow domain data to properly drive the training. 
Another way to take advantage of a large non-domain specific dataset, and a small domain specific dataset, is to use a transfer learning approach
(cf. state of the art  in Section~\ref{sec:SLU_state_of_the_art}). This means pre-training a large amount of speech data on an ASR task in order to make the model learn the input representations of the acoustic signal. Once the model is learned, training is restarted on an SLU task with other data sets designed specifically to learn concepts and intents. 
The complete process consists of 4 steps (Figure \ref{fig:transfer_learning_datasets_eng} and Table \ref{tab:intent_slot_symbols_and_asterisk}): 3 steps for the prediction of concepts and a fourth step for intent prediction:

\begin{figure}[tb]
\centering
\includegraphics[width=1\linewidth]{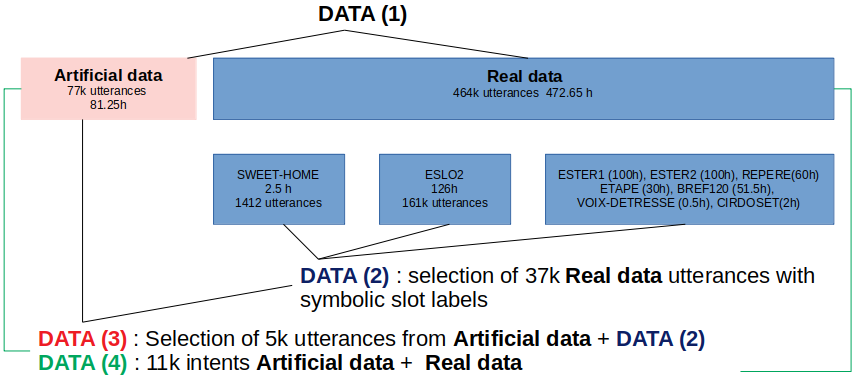}\\
\caption{Transfer learning - Concept and intent prediction} 
\label{fig:transfer_learning_datasets_eng}
\end{figure} 

\begin{enumerate}
	\item For the first step an ASR model is trained (16 epochs) on the set of real and artificial speech utterances (553.9h) (\textit{data(1)});
	\item For the second step, there is a training (12 epochs) on 37k  \textit{real} sentences of speech data, as part of \textit{data(1)}, which contains symbolic concepts specific to the home automation domain (\textit{data(2)});
	\item  The third stage is a training (9 epochs) on 3800 artificial corpus sentences whose concepts are \textit{missing} or \textit{under-represented} in \textit {real} \textit{data(2)}. 1651 statements were also taken from \textit {data(2)} containing under-represented concepts. Indeed we combined a \textit{transfer} learning and a technique of \textit{data duplication} resulting in 5451 statements (\textit {data(3)});
	\item The final step is a training (11 epochs) on intent utterances \textit{data(4)} which contains 11K statements of real and artificial speech.
\end{enumerate}

\begin{table}[th]
\caption{E2E SLU - intent and concept symbols}
\label{tab:intent_slot_symbols_and_asterisk}
\centering

\begin{tabular}{|l|}
\hline
\textbf{Concept} (data(3))\\ \hline
hestia s'il vous plaît \^{}baisser\^{} \}la lampe\} $>$de la chambre$>$ \\
\hspace{35 mm}\texttt{action} 
\hspace{2 mm}\texttt{device} 
\hspace{2 mm}\texttt{location-room} \\
(hestia please decrease light in the room) \\\hline
\textbf{Intent + Concept} (data(4))   \\
\hline
@@ hestia s'il vous plaît \^{}baisser\^{} \}la lampe\} $>$de la chambre$>$ @@\\
\hspace{0 mm}\texttt{set\_device} 
\hspace{15 mm}\texttt{action} 
\hspace{2 mm}\texttt{device} 
\hspace{2 mm}\texttt{location-room} \\
\hline
\textbf{Intent + Concept - without words outside slots } (data(4*))   \\ \hline
@@ hestia * * * \^{}baisser\^{} \}la lampe\} $>$de la chambre$>$ @@\\
\hspace{0 mm}\texttt{set\_device} 
\hspace{1 mm}\texttt{action} 
\hspace{2 mm}\texttt{device} 
\hspace{2 mm}\texttt{location-room} \\
\hline\end{tabular}
\end{table}

\begin{table}[th]
\caption{E2E SLU - intent and concept symbols}
\label{tab:eval_ESPNET_transfert_notransfert_slu}
\centering

\begin{tabular}{|l|l|l|}
\hline
\textbf{Model} & \textbf{Intent (\%)}  & \textbf{Concept (\%)} \\
 & \textbf{F1-score} & \textbf{CER}\\
\hline
\textbf{Training without transfer learning} :& & \\
Pipeline SLU & \textbf{84.21} & \emph{36.24} \\
E2E SLU & 47.31     &     51.87 \\
\hline
E2E SLU Data(3)  & - & 69.11\\
\hline
\textbf{Transfer learning E2E SLU} :& & \\
Data(1) $\rightarrow$ Data(2) & -  & 42.19\\
Data(2) $\rightarrow$ Data(3)  & - & \textbf{32.12}\\
Data(2) $\rightarrow$ Data(3)  $\rightarrow$ Data(4) & 68.13 & - \\
Data(2) $\rightarrow$ Data(3)  $\rightarrow$ Data(4*) & \emph{74.57} & - \\
\hline
\end{tabular}
\end{table}

Results are presented in Table~\ref{tab:eval_ESPNET_transfert_notransfert_slu} which shows performance results for all phases of transfer learning. The baseline result with Kaldi (\textit{Pipeline SLU}) and the E2E approach are showed for comparison (\textit{E2E SLU-small} is not used as baseline since it includes 1k from the test set). The results for the transfer learning from the ASR (\textit{Data(1)}) to the SLU task (\textit{Data(2)}), (\textit{Data(1) $\rightarrow$ Data(2)}), show lower performances than the approach with  a reduced data set. In order to verify that our results are \textit{truly} based on a  transfer learning, we compared performances based on a model trained \textit{only} on the 5k statements of \textit{Data(3)}.
Training only on \textit{Data(3)} shows its limits (CER amount to 69.11\%). On the other hand, a training on the SLU task \textit{Data(2)} that is transferred to SLU task \textit{Data(3)} (\textit{Data(2) $\rightarrow$ Data(3)}) shows its efficiency for concept prediction, performing better (CER $=$ 32.12 \%) than the baseline pipeline SLU approach (\textit{Pipeline SLU}) obtaining a CER of 36.24\%. 
These results indicate the relevance of transfer learning for E2E SLU.

For intent prediction, we continued the transfer learning principle using intent data (\textit{Data(4)}). For this intent learning task, on top of the transcriptions that are augmented with concept symbols, intent symbols were inserted (Table \ref{tab:intent_slot_symbols_and_asterisk}). Table \ref{tab:eval_ESPNET_transfert_notransfert_slu} shows that we could not outperform the sequential SLU intent prediction performances. Best transfer learning results for intent prediction were obtained using the \textit{Data(2) $\rightarrow$ Data (3) $\rightarrow$ Data(4*)} model where word tokens ``outside concepts'' have been replaced by asterisk symbols. Nevertheless the latter model outperforms the reduced model's intent prediction (\textit{ESpnet-small}).

\subsection{ASR impact on E2E SLU} \label{subsec:ASR_impact_on_E2E_SLU}

In Section \ref{subsec:E2E_SLU}, we have shown that the E2E SLU model outperforms the pipeline SLU approach, by applying transfer learning, for concept prediction in spite of ASR hypothesis transcriptions that are far from perfect. Hence, we expect a weak correlation between WER values from ASR performance and CER values from E2E SLU concept prediction, based on transfer learning. In order to verify this, we calculated the Pearson and Spearman correlation coefficients between the WER value per utterance and the CER value per utterance. Table \ref{tab:Correlations_Pearson_Spearman_WER_RAP_CER_SLU_E2E1} shows very significant correlations but which are not as strong as one would expect. Thus an improvement for an ASR task should have a positive impact on the SLU task but it is by far not sufficiently predictive of E2E SLU performances. This answers our first question by confirming that E2E approaches are indeed a way to diminish the cascade of errors effect of the ASR task on the NLU task.

\begin{table}[th]
\caption{Pearson and Spearman correlations E2E ASR - E2E SLU}
\label{tab:Correlations_Pearson_Spearman_WER_RAP_CER_SLU_E2E1}
\centering

\begin{tabular}{|l|ll|}
\hline
\textbf{Model} & \textbf{WER} (\%) & \textbf{CER} (\%)\\ 
\hline
ASR E2E & {\bf46.50}     &     - \\
E2E SLU & -     &     {\bf32.12} \\
\hline
\textbf{Correlation coef.} &      &      \\

Pearson (r) &   0.26$^{**}$   &      \\
Spearman (r$_s$)  &   0.25$^{**}$   &      \\
\hline 
\end{tabular}

$^*$ means $p<0.05$ ; $^{**}$ means $p<0.01$
\end{table}

\section{Acoustic Analysis of E2E SLU prediction} \label{sec:Acoustic_impact_on_E2E_SLU_prediction}

Unlike a sequential SLU approach, an E2E SLU approach has access to \textit{acoustic} information of the input signal. It is therefore relevant to verify whether the model exploits para-linguistic indices to infer semantic information. 

\subsection{Acoustic information impact in the E2E model}

\begin{figure}[b!t]
\begin{minipage}{0.4\linewidth}   
\scriptsize
\centering
\includegraphics[width=0.9\linewidth]{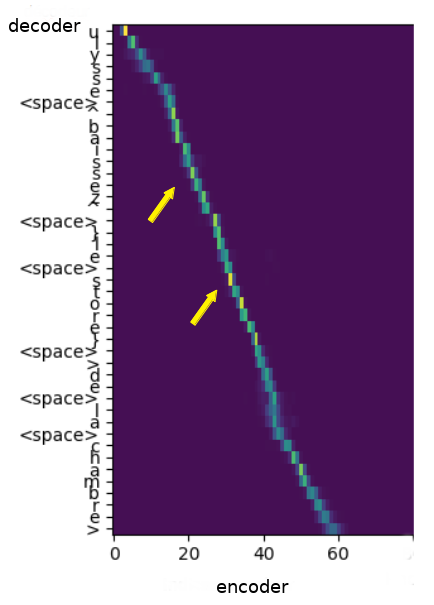} 
\caption{Utterance with attention for concept labels  
}\label{fig:attentions_concepts}
\end{minipage}\hfill
\begin{minipage}{0.5\linewidth}   
\includegraphics[width=1\linewidth]{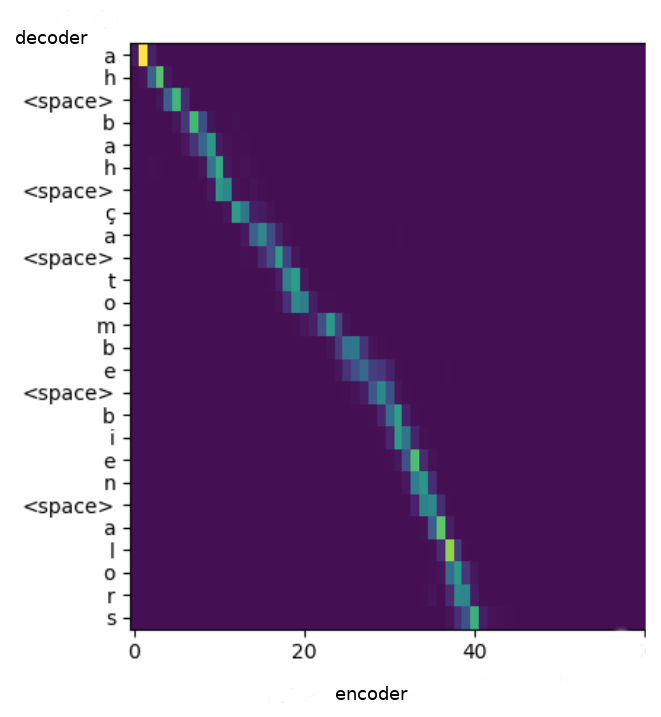} 
\caption{Utterance without concept labels}\label{fig:attentions_noconcepts}
\end{minipage}
\end{figure}

\begin{figure}[hbt]
\centering
\includegraphics[width=1\linewidth]{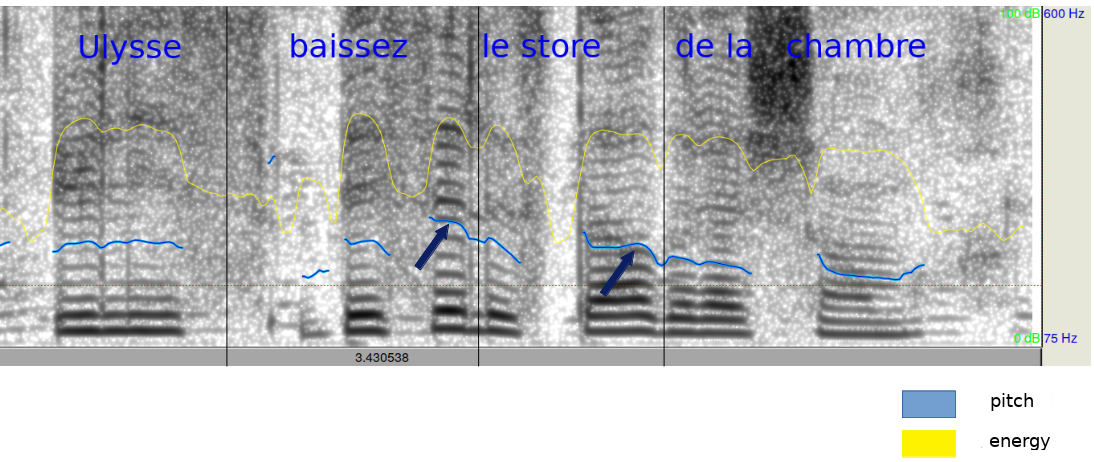}\\
\caption{Pitch and energy for the utterance ``ulysse lower the blind in the room''} 
\label{fig:pitchenergy}
\end{figure}

One way to qualitatively analyse E2E neural model is to analyse the attention map. Using ESPnet, an attention heat map was generated for the test set utterance ``ulysse baissez le store de la chambre'' ({\it ulysse lower the blind in the room}), with concept labels \verb|action| 'lower' and \verb|device| 'the blind'. The yellow arrows in Figure \ref{fig:attentions_concepts}, pointing to the lighter color areas, show increased attentions for the concept labels, especially around the hat and brace symbols that represent the concept labels \verb|action| and \verb|device| respectively. 
On top of that, pitch and energy were measured for the same utterance using \textit{Praat} \footnote{https://www.fon.hum.uva.nl/praat/}. Figure \ref{fig:pitchenergy} shows a pitch contours (blue line) that increases for the concepts.

In order to exclude that the increased attention around the symbols is caused by white-spaces ($<$space$>$), another attention heat map for a test set utterance \textit{without} concepts was generated for the utterance ``ah bah ça tombe bien alors'' ({\it ah well that's good then}) and clearly does not show any increased attention around white-spaces (Figure \ref{fig:attentions_noconcepts}). This indicates that the E2E SLU model seems to learn that the concept symbols are more important than the other character symbols. This result led us to the research question whether E2E SLU benefits from \textit{acoustic} information in order to predict concepts and intents and if correlations exist between prosody on the one hand and the prediction of slot and intent labels on the other hand.

\subsection{Background noise}

\begin{table}[th]
\caption{ASR and SLU performances - voice commands with background noise}
\label{tab:Performances_ASR_Kaldi_ESPnet_NENOISE200}
\centering
\begin{tabular}{|l|cc|cc|r|}
\hline
\textbf{Background} & \multicolumn{2}{c|}{\textbf{Pipeline SLU}}& \multicolumn{2}{c|}{\textbf{E2E SLU}}& \textbf{\#}\\
          \textbf{noise} &  \textbf{WER (\%)} &  \textbf{CER (\%)} &  \textbf{WER (\%)}&  \textbf{CER (\%)}& \textbf{utt.}\\
\hline
\textbf{All:} & & & & &\\
\hspace{0.2cm} \textbf{M\&F} &38.58&57.80&57.53&\bf 39.73&204\\

\hspace{0.2cm} \textbf{M} & 30.78 & 54.98 & 52.74 & \bf 34.05 & 152\\
\hspace{0.2cm} \textbf{F} & 58.72 & 65.06 & 69.87 & \bf 54.38  & 52\\
\hline
\textbf{Vacuum} & & & & &\\
\textbf{cleaner:} & & & & &\\
\hspace{0.2cm} \textbf{M\&F} &57.00&77.62&59.00&\bf 53.79&108\\
\hspace{0.2cm} \textbf{M} & 46.64 & 75.46 & 54.23 & \bf 47.82 & 72\\
\hspace{0.2cm} \textbf{F} & 77.75 & 81.94 & 71.31 & \bf 65.74 & 36\\
\hline
\textbf{Radio\&TV:}  & & & & & \\
\hspace{0.2cm} \textbf{M\&F}  &20.31&35.77&56.31&\bf 25.94&75\\
\hspace{0.2cm} \textbf{M}  & 18.08 & 36.64 & 53.30  & \bf 20.90 & 58\\
\hspace{0.2cm} \textbf{F}  & 27.96 & \bf 32.84 & 66.57 &  43.13 & 17\\
\hline
\textbf{Fan:} & & & & & \\
\hspace{0.2cm} \textbf{M\&F} &14.74&32.69&65.36&\bf 26.92&21\\
\hspace{0.2cm} \textbf{M} & 13.18  & \bf 24.99 & 62.84 &  38.88 & 17\\
\hspace{0.2cm} \textbf{F} & 18.27 & 49.99 & 71.03 & \bf 0 & 4\\
\hline
\end{tabular}
\end{table}

The VocADom@A4H test set was recorded in presence of background noise such as  radio, television, etc. Some of these classes of noise have high frequency components, as for example those of vacuum cleaners. We randomly selected utterances that we annotated with background noise labels until we had about 10\% (204 utterances) test set utterances with voice commands containing background noise available.

Table \ref{tab:Performances_ASR_Kaldi_ESPnet_NENOISE200} shows that E2E SLU outperforms (\textit{All}, CER) pipeline SLU, especially for utterances with high pitched vacuum cleaner background noise. Although the pipeline Kaldi ASR module outperforms ESPnet ASR in general, performance for both models is closer for utterances with vacuum cleaner background noise. This is the case for female speakers (\textit{F}) as well as for male speakers (\textit{M}). These results can be related to the results of \citep{qian2016very} according to which an E2E ASR system shows more robustness in processing noisy speech due to its layers of CNN (Convolutional Neural Network) in their architecture. However, here it is not the ASR task that benefits from it but the SLU task. It seems also that the Kaldi module exhibits much larger WER variation between gender than ESPNet. However, this behavior is reversed for the CER (since the pipeline NLU is only processing words) suggesting that the E2E SLU is influenced by acoustic features late in the prediction process to exhibit such a gender bias.

\subsection{Pitch and energy}

Pitch and energy are known to impact ASR performances. In \citep{goldwater2010words}, the prosodic characteristics that are related to an increased ASR WER error have been studied and the authors concluded that \emph{pitch} and intensity have impact at extreme values.
Pitch is a perceptive frequency-related account of a sound wave, which cannot be measured directly. 
However, two tones can be considered to have the same pitch if they share the same F0 values. Regarding speech, increasing and decreasing pitch contours help define prosody \citep{plack2005overview}. According to \citep{stehwien2016exploring} and \citep{su2018perceivable}, most of the words related to concepts also carry a pitch accent and can point to the most salient semantic information. In order to verify this, they studied the correlation between pitch variations and concepts. These studies inspired us to analyze the relationships between pitch, energy and SLU performances.

The study was performed at utterance level on the test corpus VocADom@A4H. For each utterance, the ASR performance (WER) was computed for both pipeline and E2E models and the mean F0 and mean energy were computed using the \textit{Praat} software.    

Two F0 values were computed using two filters: \begin{itemize} \item A band-pass filter between 75 and 600Hz, typically containing male and female speaker F0 values and on the other hand; \item No filter.\end{itemize}
Hence for the F0 values without filters, high pitched background noise values are also included. 
As the target of this study is the extraction of concepts and intents, the correlations were computed for the entire test set, as well as only for the 2612 test set utterances containing a voice command for comparison.

\begin{table}[th]
\caption{Pearson and Spearman correlations energy and pitch with WER of pipeline ASR (Kaldi) and E2E ASR (ESPnet)  }
\label{tab:Correlations_Pearson_Spearman_WER_RAP_CER_SLU_E2E}
\centering

\begin{tabular}{|l|cc|cc|cc|r|}
\hline
\textbf{Correlation} &  \multicolumn{2}{c|}{\textbf{Energy}}& \multicolumn{2}{c|}{\textbf{Pitch (no filter)}}& \multicolumn{2}{c|}{\textbf{Pitch(75-600Hz)}} \\
           &  \textbf{Kaldi}&  \textbf{ESPnet} &  \textbf{Kaldi} &  \textbf{ESPnet} &  \textbf{Kaldi} &  \textbf{ESPnet}\\
\hline
\hline
\textbf{full dataset}: & & & & & &\\
Pearson (r) & \textbf{-0.22$^{**}$} & -0.07$^{**}$ & -0.02 & 0.003 & 0.05$^{**}$ & 0.007  \\
Spearman (r$_s$) & -0.14$^{**}$ & -0.08$^{**}$ & -0.07$^{**}$ & -0.03$^*$ & 0.01 & 0.002 \\

\hline
\hline
\textbf{voice commands}: & & & & & &\\
\textbf{only}: & & & & & &\\

Pearson (r) & 0.04$^*$ & 0.04$^*$ & \textbf{0.23$^{**}$} & 0.18$^{**}$ & 0.09$^{**}$ &  0.06$^*$ \\
Spearman (r$_s$)&  0.05$^*$ & 0.06$^{**}$ & 0.10$^{**}$ & 0.19$^{**}$ & 0.06$^{**}$ & 0.08$^{**}$ \\
\hline
\end{tabular}
$^* p<0.05$ ; $^{**} p<0.01$
\end{table}

Table \ref{tab:Correlations_Pearson_Spearman_WER_RAP_CER_SLU_E2E} shows that Word Error Rate is significantly correlated to energy but has a real impact only on Kaldi in the full data-set case. In all other cases, in particular for voice commands the correlation is negligible.   

Regarding pitch, it is more correlated to WER for utterances with voice commands. 
This is especially the case for pitch values without filter. 
We further analyzed this effect in three steps: 
\begin{enumerate} 
\item For the voice command utterances the F0 values between 75 and 600Hz every 0.01 seconds were computed.
\item Timestamps for the word boundaries of the reference and hypothesis transcripts, with symbolic concept labels were generated, applying a forced alignment. Timestamps of F0 values on the one hand and timestamps of the word boundaries on the other hand were then aligned.
\item Finally the number of reference concept labels with the highest F0 values in the utterance that were correctly predicted in the hypothesis predictions were computed for pipeline and E2E SLU.
\end{enumerate}

The frequency list in Table \ref{tab:Freq_words_concepts_highest_F0_per_utt} shows that 3 concept labels (\texttt{device}, \texttt{action}, \texttt{location-room}) are among the 10 most frequent concepts and words with the highest F0 value per utterance. 47.79\% of all voice commands (1222/2557 voice commands) in the test set contain a concept consisting of words with the highest F0 values per utterance. It turns out that speakers, by talking to the home system, make more effort in uttering commands and thus speak with an increased intonation, which results in higher F0 values for the words belonging to concepts of the command.

\begin{table}[!ht]
\caption{Frequency of words and associated concepts with highest F0 value per utterance over 2557 voice commands of the test set}
\label{tab:Freq_words_concepts_highest_F0_per_utt}
\centering
\begin{scriptsize}
\begin{tabular}{|l|l|l|l|}
\hline
\textbf{Frequency} & \textbf{Word} & \textbf{Frequency} & \textbf{Word}\\
\hline
538&\textbf{\} (device)}&62&vocadom\\
509&\textbf{\^{} (action)}&59&est-ce\\
163&cirrus&51&hé\\
160&dis&47&hestia\\
131&ulysse&43&chanticou\\
131&\textbf{$>$ (location-room)}&39&allô\\
105&téraphim&37&que\\
84&ichéfix&35&messire\\
72&minouche&32&, (device-setting)\\
\hline
\end{tabular}
\end{scriptsize}
\end{table}

For the 1222 voice commands (reference utterances) with the highest concept pitch value, we calculated whether these concepts were well predicted by the pipeline and E2E SLU systems. As shown in Table  \ref{tab:ref_voice_command_concepts_highest_F0_in_hyp}, a slightly higher percentage of concepts are retrieved for the E2E model as compared to the SLU pipeline model. Although the difference is small this might indicate that the E2E SLU is slightly less impacted by pitch effects than the pipeline SLU.

\begin{table}[!hb]
\caption{Reference voice command concepts with highest F0 values in hypothesis transcriptions}
\label{tab:ref_voice_command_concepts_highest_F0_in_hyp}
\centering
\begin{tabular}{|l|l|}
\hline
\textbf{SLU model} & \textbf{concept ref. in hyp.(\%)} \\
\hline
Pipeline & 74.22\\
E2E & \textbf{75.00}\\
\hline
\end{tabular}
\end{table}

\subsection{Impact of MFCC and fbank features}

Another way to check the impact of pitch on ASR and SLU performance is the removal of pitch variation from the test set utterances. To this end, average values of F0 per  speaker were calculated. Using Praat, all test utterances for each speaker were resynthesized based on the resulting F0 average. As a next step, an E2E ASR model was also trained, using \textit{MFCC} features, instead of \textit{fbank} features, in order to compare the performance of ESPnet with the same acoustic features as used for Kaldi.
Table \ref{VocADom@A4H_Performances_RAP_SLU_pitch_removal}
shows that, at the level of the pipeline SLU ASR module (\textit{Pipeline ASR}), the performances of Kaldi (MFCC) on data \textit{without} pitch variation are superior to performances on data \textit{with} pitch variation. However \textit{E2E ASR} and \textit{E2E SLU} performances on data \textit{with} pitch variation are superior to utterances \textit{without} pitch variation, especially with fbank features.

Finally utterances from male (\textit{M(1)}) and female (\textit{F(1)}) speakers were evaluated separately. These utterances were compared with those of male (\textit{M(2)}) and female (\textit{F(2)}) speaker samples with a pitch \textit{above} the average per speaker. These results show that performances for (\textit{M(2)}) and (\textit{F(2)}) with \textit{deletion} of pitch variation are significantly worse than with pitch variation. This indicates that the E2E SLU is more robust to pitch variation.

\begin{table}[th]
\caption{ASR and SLU performances (\%), deletion of pitch variation}
\label{VocADom@A4H_Performances_RAP_SLU_pitch_removal}
\centering
\begin{tabular}{|l|l|cc|cc|}
\hline
\textbf{Model} & \textbf{Acoust.} &\multicolumn{2}{c|}{\textbf{No pitch var.}}& \multicolumn{2}{c|}{\textbf{Pitch var.}} \\

         & \textbf{param.} &  \textbf{WER} &  \textbf{CER } &  \textbf{WER}&  \textbf{CER} \\
\hline
\textbf{Pipeline ASR:}  &  &  &  &  & \\
\textbf{Kaldi }  & MFCC & \textbf{21.48} & - & 22.92 & - \\

\hline
\textbf{E2E ASR:}  &  &  &  &  & \\
\textbf{ESPNet ASR:} 
 & MFCC & 49.90 & - & 47.60 & - \\
 & fbank & 50.20 & - & \textbf{46.50} & - \\
\hline
\textbf{E2E SLU:}  & fbank & - & 40.02 & - & \textbf{32.12} \\
 \textbf{M(1).}   & fbank & - & 41.94 & - & 32.90 \\
 \textbf{F(1).}   & fbank & - & 36.58 & - &  30.74 \\

\textbf{F0 $>$ F0 avg.}  &  &  &  &  &  \\
 \textbf{M(2).}   & fbank & - & 53.32 & - &   42.40 \\
 \textbf{F(2).}   & fbank & - & 37.89 & - &  32.36\\
\hline
\end{tabular}
\end{table}

\section{Grammatical Analysis of E2E SLU prediction}  \label{sec:Symbolic_impact_on_E2E_SLU_prediction}

Although some pipeline approaches to SLU are able to handle OOV and syntactic variation, the E2E approach builds its own internal representation of utterances and has as final target a character string generation. This model should thus be more robust to OOV than classical ASR/NLU modules. To assess the capability of the E2E SLU model to handle better  grammatical variations than a pipeline SLU system, we generated new input stimuli for which we controlled the \textit{linguistic variation}, in particular at the \textit{lexical} and \textit{syntactic} levels. This is detailed in the two following sections.

\subsection{Out of vocabulary words (OOV)}

In order to measure the impact of an increased OOV rate, we gradually replaced the test set vocabulary of some specific concepts with words that did not appear in the training set. To measure an increasing difficulty this experiment was performed in 4 steps~: \begin{enumerate} \item Step 1: {\tt action} and {\tt device-setting}, \item Step 2:  Step 1 and {\tt device} , \item Step 3:  Step 2  and {\tt location}, \item Step 4:  Step 3 and {\tt key-words}. \end{enumerate}
The following example shows a voice command (``vocadom turn on the kettle'') with symbolic intent and concepts \textit{before} (1) and \textit{after} (2) insertion of OOV words (Step 4):

\begin{small}
\begin{lstlisting}
(1)  @ ah vocadom euh ^allume^ }la bouilloire} @ 
(ah vocadom uh turn on the kettle)

(2)  @ ah ursule euh ^enclenche^ }la bouillotte} @
(ah ursule uh switch on the kettle)
\end{lstlisting}
\end{small}

Table \ref{tab:OOV_setup} shows that substituted words in step 4 represent 26.15 \% of the total number of word tokens (31k) and 3.48 \% of the total number (1462) of word types (vocabulary).
\begin{table}[th]
\caption{Vocadom@A4H - ratio OOV total words}
\label{tab:OOV_setup}
\centering
\begin{scriptsize}
\begin{tabular}{|l|c|c|c|c|}
\hline
\textbf{Substitutions}  & \textbf{\#Word}  & \textbf{\#Words} & \textbf{(\%) Word} & \textbf{(\%) Total} \\ 
  & \textbf{Type}  &  & \textbf{Type} & \textbf{Words} \\ \hline
Step 1 & 22 & 1785 & 1.50 & 5.72 \\
Step 2  & 34 & 4276 & 2.32 & 13.70 \\
Step 3  & 41 & 5516 & 2.80 & 17.68 \\
Step 4  & 51 & 8160 & 3.48 & 26.15 \\
\hline
\end{tabular}
\end{scriptsize}
\end{table}

Once the test set has been altered with OOV words, the speech utterances were generated with the same TTS tool as used for the artificial corpus generation in Section~\ref{sec:synthdata}. The resulting utterances were then fed to the two SLU models. However, the E2E SLU was trained on data containing artificial speech, which was not the case for the pipeline SLU approach. Hence, for a fair comparison, the resulting E2E ASR model was used as ASR front-end for the pipeline SLU.

\begin{table}[!ht]
\caption{Impact of OOV on SLU performances (\%)}
\label{tab:Evaluation_VocADom@A4H_SLU_pipeline_OOV}
\centering
\begin{tabular}{|l|cc|cc|ccc|}
\hline
\textbf{Model}  &\multicolumn{4}{c|}{\textbf{Pipeline}}&\multicolumn{3}{c|}{\textbf{E2E}} \\
  &\multicolumn{2}{c|}{\textbf{NLU}}& \multicolumn{2}{c|}{\textbf{ASR+NLU}} & \multicolumn{3}{c|}{\textbf{SLU}} \\
          &  \textbf{CER} &  \textbf{F1} &  \textbf{CER}&  \textbf{F1}&  \textbf{WER} &  \textbf{CER}&  \textbf{F1}\\
\hline

Compl. real & 33.78 & 85.51 & 36.24 & 84.21& 46.50  & 32.12  & 74.57\\
\hline
Compl. synth.   & - & - & 37.07 & 83.34 & 39.30 & 25.00 &  53.70 \\

\hline
\textbf{OOV:}  &  &  &  & &  &  & \\
Step 1 &   37.75 & 81.50 & 45.43 & 79.56& 44.00 & 30.75 & 50.39\\
Step 2 &   53.77 & 72.39 & 62.03  & 72.48 & 53.20 & 46.75 & 50.26\\
Step 3  &   63.01 & 69.58 & 68.07 & 70.29 & 52.50 & 50.89 & 51.59\\
Step 4   &   90.45 & 63.66 & 86.44 & 65.03 & 55.90 & 58.80 & 51.43 \\
\hline
\textbf{Diff.} & 56.67 & 21.85 & 49.37 & 18.31& \textbf{16.6} & \textbf{33.8}  & \textbf{2.27} \\
\hline
\end{tabular}
\end{table}

Table \ref{tab:Evaluation_VocADom@A4H_SLU_pipeline_OOV} shows ASR and SLU performances according to the rate of OOV words. The complete \textit{Compl. real.} line corresponds to the original Vocadom@A4H test set while the \textit{Compl. synth.} lines corresponds to the test set of which speech has been synthesized through TTS. Using the E2E ASR for the pipeline SLU did increase the CER (ASR+NLU column) but had a small impact on the intent prediction. Overall, for all models performances for concept (\textit{CER}) and intent  prediction (\textit{F1}-score) 
deteriorate with increased OOV rates. The differences (\textit{Diff.}) between \textit{Compl. synth.} and \textit{Step 4} are much smaller for the E2E model than for the pipeline SLU model for both concept and intent prediction. This would suggest that the E2E model is more robust to OOV words. 

It must be noticed that the intent prediction of the E2E model, decreases dramatically with synthetic speech. This is due to an increased error rate for the {\tt None} intent that consisted only of real speech in the training data whereas we used synthetic evaluation data for OOV impact evaluation. This is another evidence that acoustic features play a role up to the higher decision stages of the E2E model.

\subsection{Syntactic variation}

In this section, we measure the robustness of both SLU models for syntactic variability, predicting concepts and intents on test data with progressive syntactic variability in two steps:\begin{enumerate}
    \item Step 1, 32 verbs belonging to the {\tt action} concept were replaced by more complex syntactic constructions;
    \item Step 2, substitutions in Step 1 have been augmented by disfluencies surrounding the words of 18 labeled concepts of {\tt device}.
\end{enumerate}

The following example ('Vocadom turn on the kettle') shows a voice command containing an intent and symbolic concepts from the test set \textit{before} (1) and \textit{after} (2) insertion of more complex syntactic constructions and disfluencies (Step 2):

\begin{small}
\begin{lstlisting}
(1)  @ vocadom euh ^allume^ }la bouilloire} @
(vocadom uh turn on the kettle)
(2)  @ vocadom euh pourrais-tu ^allumer^ la la }bouilloire} @ 
(vocadom uh could you turn on the the kettle)
\end{lstlisting}
\end{small}

We also generated text-to-speech based on the resulting modified test sets, that we evaluated in the same way as for the OOV words test set up, as outlined in previous section.

\begin{table}[!ht]
\caption{Impact of syntactic variation on SLU performances (\%)}
\label{tab:Evaluation_VocADom@A4H_SLU_E2E_OOV}
\centering
\begin{tabular}{|l|cc|cc|ccc|}
\hline
\textbf{Model}  &\multicolumn{4}{c|}{\textbf{Pipeline}}&\multicolumn{3}{c|}{\textbf{E2E}} \\
  &\multicolumn{2}{c|}{\textbf{NLU}}& \multicolumn{2}{c|}{\textbf{ASR+NLU}} & \multicolumn{3}{c|}{\textbf{SLU}} \\
          &  \textbf{CER} &  \textbf{F1} &  \textbf{CER}&  \textbf{F1}&  \textbf{WER} &  \textbf{CER}&  \textbf{F1}\\
\hline

Compl. real & 33.78 & 85.51 & 36.24 & 84.21& 46.50  & 32.12  & 74.57\\
\hline
Compl. synth.   & - & - & 37.07 & 83.34 & 39.30 & 25.00 &  53.70 \\

\hline
\textbf{Synt. var.:}     &  &  &  & &  &  &   \\
Step 1     & 38.41 & 81.06 & 50.40 & 77.45  & 44.40 & 16.29 &  52.59\\
Step 2     & 38.34 & 81.19 &  52.75 & 76.36  & 50.90 & 22.07 &  49.09\\
\hline

\textbf{Diff.} & 4.56 & 4.32 & 15.68 & 6.98& \textbf{11.60} & \textbf{2.93} & 4.61\\
\hline
\end{tabular}
\end{table}

Table~\ref{tab:Evaluation_VocADom@A4H_SLU_E2E_OOV}
shows that differences (\textit{Diff.}) between performances of concept and intent prediction for test data with the complete original syntactic structure \textit{Compl. synth.} on the one hand and test utterances with modified syntactic structure \textit{Synt. var., Step 2} on the other hand, are smaller for the E2E model than for the sequential SLU model. This again indicates a greater robustness of the E2E model to cope with increased syntactic variation.
Table \ref{tab:Evaluation_VocADom@A4H_SLU_E2E_OOV} also shows that the performance of the E2E model for concept prediction improves with a more complex syntax. This may be due to an average sentence length of 15 words for the artificial corpus utterances, while the average sentence length for the (original) test set utterances is only 5. The increased syntactic variation, also increases the length of the evaluation utterances which consequently approaches the average length of the the artificial corpus utterances.

\begin{table}[th]
\caption{Impact of utterances length in the real test set case}
\label{tab:SLU_performances_on_longest_utterances}
\centering
\begin{tabular}{|l|cc|cc|r|}
\hline
\textbf{Test set} & \multicolumn{2}{c|}{\textbf{Pipeline SLU}}&
\multicolumn{2}{c|}{\textbf{E2E SLU}}\\

\textbf{} &  \textbf{Concept} &  \textbf{Intent} &  \textbf{Concept}&  \textbf{Intent}\\
\textbf{} &  \textbf{CER (\%)} &  \textbf{F1 (\%)} &  \textbf{CER (\%)}&  \textbf{F1 (\%)}\\

\hline
\hline

\textbf{Compl. real}  & 36.24  & 84.21 &  32.12 &   74.57\\
6747 utterances & & & &\\
\hline
\textbf{Compl. real $>$ 7 words} & 38.62 & 80.01 & 33.18 &  68.23\\
1461 utterances &  & &  &  \\
\hline 
\textbf{Diff.} & \bf 2.38 & 4.2 & {\bf1.06} & 6.34\\
\hline
\end{tabular}
\end{table}

To test whether such robustness of the E2E SLU model is observable on the \textit{real} test data (and not only on synthetic data), we extracted the more complex utterances from the test set based on their length. Since 
the average sentence length of the test set is seven words, 1461 utterances of the original test set (over 6747 total utterances) were labelled more complex. Although higher number of words does not necessarily means more complex sentences, this is nevertheless correlated in the voice command case. The performances are shown on Table~\ref{tab:SLU_performances_on_longest_utterances}. The first row recall the results on the whole test set while the second row shows performances computed on long utterances only. As with the synthetic dataset, the E2E SLU concept prediction errors increases far less with the length of sentences than the pipeline SLU (Table \ref{tab:SLU_performances_on_longest_utterances}, Diff.). Furthermor, similarly to Table~\ref{tab:eval_ESPNET_transfert_notransfert_slu}, we can see that the E2E SLU could not outperform the pipeline SLU intent prediction performances.

The reader can find a summary of the experiments measuring the acoustic and grammatical impact on the E2E SLU model compared with the pipeline SLU model in Table~\ref{tab:Evaluation_aperçu_results}.

\section{Discussion}\label{sec:Discussion}

To deal with the lack of data our strategy consisted in generating artificial utterances for voice commands. To deal with the bottleneck of distance between real speech test data and artificial speech training data, we applied a \textit{transfer learning} approach. A initial model was pre-trained on large out-of-domain data but composed of real speech data, and then this model was fine-tuned on the  artificial but in-domain speech. It allowed us to take advantage of a large non-domain specific data set, and a small domain specific data set. This approach outperformed the pipeline SLU for concept prediction. Our data augmentation technique is close to \citep{li2018training,lugosch2019using}, who reported better concept prediction performances with a real speech model augmented with artificial speech than with an acoustic model only trained on real speech. \cite{li2018training} reported optimal performances for their E2E ASR model using an acoustic model trained on 50\% synthetic speech data and 50\% real speech.
On the other hand, \textit{intent} prediction did not sufficiently benefit from transfer learning. A possible explanation is that ESPnet, being an ASR tool, functions at a level that is too local to be able to perform the global abstraction required for intent prediction. 

Regarding the analysis of performances, for the Pipeline SLU model, performance differences between the NLU module and the complete sequential SLU system often remain high, despite the strategies used to reduce them. Since the current state of the art indicates that good SLU performance requires good ASR performance, we compared the pipeline and E2E SLU approaches on an ASR task learned on equivalent data. Our experiments show that the WER of the E2E ASR is significantly higher than the one of the pipeline ASR module. However, the E2E SLU approach shows better SLU performance for concept prediction, using the same tool (ESPnet) as E2E ASR.  On top of that, our correlation tests in Section \ref{subsec:ASR_impact_on_E2E_SLU} showed that perfect ASR is not necessary to obtain good E2E SLU performance. It is however essential in the case of a pipeline approach as we have demonstrated in \citep{desot2019towards} for intent prediction and in \citep{desot2019slu} for concept prediction. \textbf{This answers our first question and confirm the state-of-the-art: the E2E model reduces the cascade of errors effect.}

The E2E approach infers concepts and intents conveyed by an utterance directly from the acoustic signal.  Our experiments in Section~\ref{sec:Acoustic_impact_on_E2E_SLU_prediction} reveal that prosodic information allows the model to point to the most important semantic information. There are indications that the higher pitch values improve the performance of the E2E SLU approach that turns out to be more robust to noisy speech as compared to the pipeline model. This indicates that the convolutional network of the E2E SLU model seems to benefit more from correlated and richer filter-bank features than from MFCC features, used by the pipeline ASR model. \textbf{This answer our second research question by showing that the E2E SLU model uses prosodic information to infer concepts and is able to learn a feature representation more robust to noise than the pipeline model.} 

As our target users are senior adults who tend to deviate easily from a fixed grammar of voice commands, we tested the SLU approaches with increased amount of OOV words and inserted more syntactic variation in the test corpus utterances. In these two cases, the E2E SLU model proved to be more robust than the pipeline approach while both model have been exposed to the same data to learn the concepts to extract (same NLU training set). \textbf{This answers our third research question. The E2E SLU model is indeed more robust to syntactic variation than the pipeline model.}

\section{Conclusion and future work}\label{sec:Conclusion}

Our answer to the bottleneck of the cascade of errors effect between the ASR and NLU modules of a pipeline SLU model, was an E2E SLU model that extracts intents and concepts directly from the signal. This approach based on deep neural networks allowed us to avoid the \textit{cascade of errors} by performing a joint learning of these two tasks in one and the same model. By comparing our E2E SLU approach with a pipeline baseline approach, composed of a state of the art ASR system and an NLU module, learned on data, specific to the home automation field, we were able to show that the E2E SLU approach gives best performance in terms of concept prediction. A possible solution to improve E2E SLU intent prediction is adding a decoder to the ESPnet architecture in order to train and predict concepts and intents jointly. A similar multi-task learning has already been applied for NLU in \citep{Liu2016}.

We can confirm one of the conclusions of the study of \cite{stehwien2016exploring} and \cite{su2018perceivable} that prosodic information can point to the most important semantic information. We have shown that the E2E SLU model exploits prosodic information which favours its performance in predicting intents and in particular concepts. On top of that the E2E SLU model shows more robustness as compared to the pipeline approach for processing target users syntactic variation and an increased OOV rate.

A transfer learning allowed us to decrease the \textit{acoustic} distance between the artificial speech of the training data and the real speech of the test set. To further decrease this distance, speech synthesis based on a larger number of voices could be generated and added to the training data. Another possibility is training a neural speech synthesis model such as \textit{Tacotron} \citep{wang2017tacotron, li2018training} for one or more speakers of the real SWEET-HOME corpus. The resulting model could then  be used to generate new synthetic utterances that would be closer to the reference corpus.

One of the strengths of the E2E SLU approach is its processing of noisy speech data, which makes this approach very suitable in a realistic smart home situation where voice commands must be extracted from utterances with various background noises. In addition to background noise, residents are far away from microphones. This leads to distorted acoustic signals by reverberation depending on the acoustics of the room. To better assess and understand the performance of the E2E SLU system in such a realistic situation, as a next step, the micro-distant recordings version of the test corpus should be tested. These are the recordings made by the four antennas of 4 microphones integrated into the ceiling of the Amiqual4Home smart home. This could be solved by augmenting our training data with \textit{room impulse response} (RIR) data. However, the acquisition of real RIR data is not trivial. \cite{ko2017study} show that acoustic models trained on simulated RIR data are competitive with real RIR data. Therefore, a technique for increasing our training data with simulated RIR data could be explored.

\section*{Acknowledgements}\label{sec:Acknowledgements }
This work is part of the VOCADOM project founded by the French National Research Agency (Agence Nationale de laRecherche) / ANR-16-CE33-0006. It was also partially supported by MIAI@Grenoble-Alpes (ANR-19-P3IA-0003).

\bibliographystyle{model5-names}
\bibliography{references}

\begin{appendix}
\newpage

\section{Some examples of variants of a window opening command}

\begin{tabular}{|l|p{6cm}p{6cm}|}

\hline
 & French & English translation\\
\hline
1 & Ouvre la fenêtre & Open the window\\
2 & Ouvre la fenêtre s'il vous plaît & Open the window please\\
3 & Est-ce que tu peux ouvrir la fenêtre? & Can you open the window?\\
4 & Est-ce que tu peux ouvrir la fenêtre s'il vous plaît? & Can you open the window please?\\
5 & Je veux que tu ouvres la fenêtre & I want you to open the window\\
\hline
\end{tabular}
\section{Overview of measures of the acoustic and grammatical impact on the SLU models}
\begingroup 
\let\clearpage\relax

\begin{sidewaystable}[ht]

\caption{Overview of Pipeline and E2E SLU evaluation results on VocADom@A4H test data}
\label{tab:Evaluation_aperçu_results}
\centering
\begin{footnotesize}
\begin{tabular}{|l|cccc|ccccc|ccccc|cc|}

\hline
\textbf{Analysis level:}  &\multicolumn{9}{c|}{\textbf{Acoustic}} &    \multicolumn{7}{c|}{\textbf{Symbolic}} \\
\hline
 &\multicolumn{7}{c|}{\textbf{Source}} & \multicolumn{2}{c|}{\textbf{Prosody}}  & \multicolumn{5}{c|}{\textbf{Lexical}} & \multicolumn{2}{c|}{\textbf{Syntax}}\\
\hline
\textbf{Analysis:}  &\multicolumn{4}{c|}{\textbf{Noise}} & \multicolumn{3}{c|}{\textbf{Gender}} & \multicolumn{2}{c|}{\textbf{Avg. F0}} &  \multicolumn{5}{c|}{\textbf{OOV}} & \multicolumn{2}{c|}{\textbf{Variation}}  \\
\hline

\textbf{Results:} & \textbf{V} & \textbf{RT} & \textbf{F} & \multicolumn{1}{c|}{\textbf{A}} &  \textbf{M}  & \textbf{F} & \multicolumn{1}{c|}{\textbf{M\&F}}&  \textbf{MFCC} & \textbf{fbank} &  \textbf{0} & \textbf{1}  & \textbf{2}  & \textbf{3} & \textbf{4} & \textbf{1} & \textbf{2}\\
\hline
\textbf{Pipeline SLU :} &  &  &  &  & &    & \multicolumn{1}{c|}{} &   &  \multicolumn{1}{c|}{} &     &   &  &  &  &  &\\
WER & 57.00 & 20.31 & 14.74 & \multicolumn{1}{c|}{38.58} & 22.25 &  24.12 &  \multicolumn{1}{c|}{22.92}  &  21.48 & \multicolumn{1}{c|}{-}  &  -  & -  & - & - & - & - & -\\
Concept Error Rate &  77.62 & 35.77 &  32.69 & \multicolumn{1}{c|}{ 57.80} & 37.56 & 33.77 &  \multicolumn{1}{c|}{36.24} &  - & \multicolumn{1}{c|}{-}  & 37.07 & 45.43  & 62.03  & 68.07  & 86.44  & 50.40 & 52.75\\
Intent F-measure & - & - & - & \multicolumn{1}{c|}{-} & 83.64 & 86.48 &  \multicolumn{1}{c|}{84.21} &  - &  \multicolumn{1}{c|}{-} &   83.34 &  79.56   & 72.48  & 70.29 &  65.03 &  77.45 &  76.36 \\

\hline  
\textbf{E2E SLU :} &  &  &  &  & &    & \multicolumn{1}{c|}{} &   &  \multicolumn{1}{c|}{} &     &   &  &  &  &  &\\

WER & 59.00 & 56.31 & 65.36 & \multicolumn{1}{c|}{57.53} & 46.90 & 44.00 &  \multicolumn{1}{c|}{46.50} & 49.90  &  \multicolumn{1}{c|}{50.20}  & 39.30 & 44.00   & 53.20 & 52.50 & 55.90  & 44.40 & 50.90 \\
Concept Error Rate &  53.79 &  25.94 &  26.92  & \multicolumn{1}{c|}{39.73} & 32.90 & 30.71 & \multicolumn{1}{c|}{32.12} & -  &  \multicolumn{1}{c|}{40.02} &  25.00  &  30.75 & 46.75 & 50.89 & 58.80  & 16.29 & 22.07 \\
Intent F-measure & -  & -  &  - & \multicolumn{1}{c|}{-} &  74.30 & 75.04  & \multicolumn{1}{c|}{74.57} &  - & \multicolumn{1}{c|}{-}  &  53.70  & 50.39  & 50.26 & 51.59  & 51.43 & 52.59 & 49.09\\
\hline
\end{tabular}
\end{footnotesize}

\caption{Explication of terminology used in table \ref{tab:Evaluation_aperçu_results} summarizing SLU performances}
\label{tab:explic_aperçu_evaluation}
\setlength{\tabcolsep}{10pt}
\centering
\begin{tabular}{|l|l|l|}
\hline
\textbf{Analysis} & & \textbf{Explication} \\
\hline
Noise  & V & Vacuum cleaner\\
       & RT & Radio-Television\\
       & F & Fan\\
       & A &  All background noises\\
       \hline
Gender & M & Masculine\\
      & F & Feminine \\
      & M\&F & Masculine and Feminine\\
\hline      
Avg.F0     & mfcc, fbank  & Average F0 value per speaker, mfcc, fbank\\
\hline
Lexical OOV              &  0 &  Artificial Speech synthesis, complete vocabulary \\ 
                 &  1, 2, 3, 4 &  Out of vocabulary, steps 1, 2, 3, 4\\
\hline                 
Syntactic variation              &  1, 2 &   Steps 1, 2\\         
\hline
\end{tabular}
\end{sidewaystable}
\newpage

\end{appendix}
\endgroup 
\end{document}